\documentclass{article}
\usepackage{iclr2027_conference,times}
\usepackage{graphicx}
\usepackage{url}
\usepackage{natbib}
\usepackage{caption}
\usepackage{booktabs}
\usepackage{amsmath}
\usepackage{amssymb}
\usepackage{amsfonts}
\usepackage{array}
\usepackage{algorithm}
\usepackage{algorithmic}

\usepackage{hyperref}

\title{DEFlow: Deep Equilibrium Flow Maximum-Entropy Reinforcement Learning}

\author{
Ruslan Ageev$^{1,*}$,
Huaze Tang$^{1,*}$,
Runqing Miao$^{1,2,*}$,
Jiarong Li$^{2,\dagger}$,
Wenbo Ding$^{1,\dagger}$\\
$^{1}$ Tsinghua University\\
$^{2}$ Peng Cheng Laboratory\\
$^{*}$ Equal contribution.\\
$^{\dagger}$ Co-corresponding authors.
}

\iclrfinalcopy
\begin{document}
\raggedbottom

\maketitle

\begin{abstract}
Expressive flow policies can represent multiple high-value actions, but sampling them with few integration steps while retaining tractable action densities remains challenging. We introduce DEFlow, a deep equilibrium flow policy that replaces explicit Euler sampling with implicit Euler integration and solves all intermediate actions jointly as an equilibrium system. We combine implicit differentiation with a discrete change-of-variables formula for the final-action density, enabling off-policy training within soft actor--critic without backpropagating through solver iterations. DEFlow achieves an average relative improvement of approximately $10\%$ in mean successful coverage over the strongest baseline across four MultiGoal docking variants. It also achieves the highest mean time-averaged return on four of five manipulation tasks, while remaining broadly competitive across five MuJoCo tasks. These results support implicit flow policies for tasks requiring both diversity and precision, at the cost of additional nonlinear and linear solves.\footnote{Code available at \url{https://anonymous.4open.science/r/deflow-81BE/README.md}.}
\end{abstract}

\section{Introduction}

Soft actor--critic (SAC) is a strong baseline for continuous control. It combines data-efficient off-policy learning with entropy-driven exploration \citep{haarnoja2018sac,haarnoja2018sacapps}. Its standard actor, however, is a single tanh-squashed diagonal Gaussian. This distribution is easy to sample from and has a closed-form density, but it cannot easily represent several distinct high-value actions at the same state \citep{haarnoja2017sql}. Diffusion and flow-matching policies offer a more expressive alternative. They generate actions by transforming simple noise through multiple learned steps and can represent complex, multimodal distributions \citep{wang2023diffusionql,lipman2023flowmatching}. The challenge is to gain this expressivity while keeping SAC training practical. Sampling and actor differentiation now involve a multi-step rollout, and the final-action log-density required by the entropy objective may be hard to obtain.

Existing off-policy methods resolve this tension by design compromise. Some give up the maximum-entropy objective altogether: Flow Q-Learning distills a multi-step generator into a one-step actor \citep{park2025fql}, consistency policies replace the long sampler with one- or few-step generation \citep{ding2024consistencypolicy}, and Diffusion-QL combines diffusion behavior cloning with value maximization without evaluating the policy likelihood \citep{wang2023diffusionql}. Q-score matching (QSM) \citep{psenka2024qscore} directly matches the policy score to the critic's action-gradient and avoids backpropagating through the complete sampler. SAC-Flow \citep{zhang2025sacflow}, the direct parent of this work, treats its explicit $K$-step flow rollout as a sequential model and stabilizes the recurrence with a gated velocity field. DIME retains a multi-step denoising policy and trains its transitions through approximate inference toward a $Q$-defined energy-based target \citep{celik2025dime}. These methods make expressive generative actors trainable off-policy, but leave a separate numerical question: how should a multi-step policy be sampled when its learned velocity field becomes difficult to integrate with a small number of explicit steps? We investigate implicit-Euler sampling as an alternative.

We introduce DEFlow, a flow policy built around implicit-Euler sampling. We retain SAC-Flow's gated velocity field and multi-step actor, but evaluate each velocity at the unknown next action, replacing the explicit Euler recurrence with implicit-Euler equations \citep{hairer1996solving}. This integration rule is well suited to stiff dynamics, making it a natural alternative when a learned field is poorly resolved by a few explicit steps. To implement the sampler, we collect all $T$ intermediate actions into one small deep-equilibrium system and solve for them jointly \mbox{\citep{bai2019deq,walker2011anderson}}. The forward pass uses a root solver to obtain the trajectory, while the backward pass is formulated through implicit differentiation rather than backpropagation through the solver iterations. Under the stated regularity assumptions, an exact equilibrium also admits a final-action log-density through discrete change of variables. The implementation evaluates the density and implicit gradient along the numerically obtained trajectory.

In summary, our contributions are threefold:
\begin{itemize}
    \item \textbf{A joint equilibrium formulation of flow policies.} We formulate the full implicit-Euler action trajectory as a single deep equilibrium system, providing an alternative to explicit rollouts for expressive policies.
    \item \textbf{Maximum-entropy policy learning.} We train DEFlow within SAC using implicit gradients and a final-action density formula, exact at equilibrium under the stated assumptions.
    \item \textbf{Comprehensive empirical evaluation.} We study DEFlow across precision docking, manipulation, and MuJoCo benchmarks, with matched explicit-Euler baselines, step-count ablations, and numerical density verification. The experiments characterize both the strengths and the limitations of implicit flow sampling.
\end{itemize}

\section{Background and Related Work}
\label{sec:related}

\subsection{Maximum-Entropy Reinforcement Learning}

We consider an infinite-horizon Markov decision process with state $s\in\mathcal{S}$,
action $a\in\mathcal{A}\subseteq\mathbb{R}^{d_a}$, transition kernel $p(s'\mid s,a)$,
reward $r(s,a)$, and discount $\gamma\in(0,1)$, where $d_a$ is the action dimension. Soft actor--critic (SAC) trains
a policy $\pi_\theta$ off-policy on transitions drawn from a replay buffer $\mathcal{D}$
to maximize the maximum-entropy return
$J(\pi_\theta)=\mathbb{E}_{\tau\sim\pi_\theta}\big[\sum_t\gamma^t\big(r(s_t,a_t) + \alpha\,\mathcal{H}(\pi_\theta(\cdot\mid s_t))\big)\big]$,
where the temperature $\alpha>0$ trades reward against entropy \citep{haarnoja2018sac,haarnoja2018sacapps}.
With twin critics $Q_{\phi_1},Q_{\phi_2}$, the actor balances expected value
against the log-density of its sampled actions. The same log-density enters
the soft target for the critics. Section~\ref{sec:recalibration} gives the
actor and critic objectives used by DEFlow.

SAC provides the actor--critic framework used in this work. Its standard
actor parameterizes a state-conditioned Gaussian distribution whose samples
are squashed by $\tanh$ to satisfy the action bounds
\citep{haarnoja2018sacapps}. The underlying Gaussian is unimodal,
which limits the policy's ability to capture
complex, multimodal action distributions.

\subsection{Flow and Diffusion Policies}

A flow policy is the expressive actor we adopt. Rather than a single Gaussian
draw, it turns base noise $z\sim p_0$ into an action by transporting it through
a learned velocity field. The state-conditioned transport follows
$\dot a(t)=v_\theta(a(t),s,t)$, starting from $a(0)=z$ for $t\in[0,1]$.
A numerical integrator turns this motion into $T$ discrete steps; the
endpoint is then squashed to keep the action bounded. Flow matching
\citep{lipman2023flowmatching} provides a way to learn such fields from data.
By transporting samples from a simple base distribution, a flow can
represent complex action distributions with multiple distinct modes.

Representing a richer action distribution, however, does not by itself make
the policy easy to optimize with SAC. The standard actor update propagates
gradients from the critic through the sampled action to the policy
parameters. Flow policies admit this differentiation, but direct
optimization requires propagating the learning signal through the entire
sequence of sampling steps. This can be computationally expensive and can
lead to vanishing or exploding gradients, making stable policy improvement
a separate challenge from learning to reproduce an action distribution
\citep{park2025fql,zhang2025sacflow}.

Our main comparisons concern expressive policies learned through online interaction.
SAC-Flow \citep{zhang2025sacflow}, our direct predecessor, combines a gated velocity field
with explicit Euler integration and a surrogate entropy objective within SAC.
DIME \citep{celik2025dime} instead uses a variational maximum-entropy bound, while
Q-score matching (QSM) \citep{psenka2024qscore} learns the policy score from the critic's
action-gradient without evaluating a density.

\subsection{Implicit Models and Differentiation}

\emph{Deep equilibrium} (DEQ) models are implicit-layer networks
\citep{bai2019deq}: rather than stack many explicit layers, a DEQ defines its
output as the fixed point $Z^\star=g_\theta(Z^\star)$ of a single weight-tied
map and computes it with a root-finder. When $g_\theta$ is a contraction,
the iteration $Z^{k+1}=g_\theta(Z^k)$ converges at a linear rate bounded by
its Lipschitz constant. Anderson acceleration \citep{walker2011anderson}
uses a history of iterates and residuals to accelerate this solve.

The gradient follows from the solved equilibrium, without differentiating
through the solver iterations. For a continuously differentiable map with
nonsingular $I-\partial_Z g_\theta(Z^\star)$, implicit differentiation turns
the nonlinear forward problem into a linear system in the backward pass.
Standard DEQ implementations solve the adjoint system iteratively.
Jacobian regularization can improve the convergence of these solves
\citep{bai2021jacobian}. A complementary perspective treats backpropagation
as block-triangular back substitution \citep{edelman2024backslash}.
We use the equilibrium formulation and implicit differentiation to construct
DEFlow in Section~\ref{sec:method}.
\section{The Finite-Step Sampling Problem}
\label{sec:problem}

At a fixed state, the maximum-entropy optimum~\citep{haarnoja2017sql,haarnoja2018sacapps} is
\begin{equation}
  \pi^\star(a\mid s) \propto \exp\!\left(Q^\star(s,a)/\alpha\right).
  \label{eq:maxent-policy-problem}
\end{equation}
It can therefore contain several narrow modes corresponding to different valid
actions. The tanh-Gaussian actor used by SAC is unimodal before squashing,
whereas a flow policy can represent these modes if its finite-step sampler
preserves them.

\paragraph{A finite inference budget.}
At a fixed state, explicit and implicit Euler use
\begin{align}
  a_{k+1}^{\mathrm{exp}} &= a_k+\Delta t\,v_\theta(a_k,s,t_k),
  \\[-2pt]
  a_{k+1}^{\mathrm{imp}} &= a_k+\Delta t\,v_\theta(a_{k+1},s,t_{k+1}) .
  \label{eq:problem-integrators}
\end{align}
where $t_k=k\Delta t$ and $\Delta t=1/T$. Real-time control limits latency and field evaluations, so the
usable $T$ may be small and $\Delta t$ correspondingly large. Solving an implicit
step may require multiple field evaluations, increasing its computational cost.

\paragraph{Three distinct numerical effects.}
Coarse discretization affects (i) the endpoint, (ii) the step map, and (iii)
the policy gradient. For the one-step maps $F_k\colon a_k\mapsto a_{k+1}$ and
$J_k=D_av_\theta(a_k,s,t_k)$, the step Jacobians are
\begin{equation}
  D_aF_k^{\mathrm{exp}}=I+\Delta t\,J_k,
  \qquad
  D_aF_k^{\mathrm{imp}}=\left(I-\Delta t\,J_{k+1}\right)^{-1} .
  \label{eq:problem-step-jacobians}
\end{equation}
so the two schemes can have different conditioning on the same field. In the
fixed-state mixture analysis of Appendix~\ref{app:finite-step}, the exact field
is smooth, but its largest derivative grows at least like $m^2\sigma^{-3}$ as
the mode width $\sigma$ decreases. Writing $\pi_{\Delta t}^\star$ for the law induced
by $T$ explicit Euler steps of the exact field and $\pi^\star$ for the two-mode
surrogate target of that appendix, explicit Euler incurs
\begin{equation}
  \Delta t\ge 2\sqrt{2}\,\sigma
  \quad\Longrightarrow\quad
  W_2(\pi_{\Delta t}^\star,\pi^\star)\ge \frac{\sigma}{2\sqrt{2}},
  \label{eq:problem-sampling-bound}
\end{equation}
and at $T=2$ its error tends to a positive limit as $\sigma\to0$
(Eq.~\eqref{eq:app-two-step-error}). The same analysis bounds the discrete
gradient at no more than a fraction $3(\sigma/\Delta t)^2$ of the exact one for a
specified Gaussian subfamily and terminal loss, so it nearly vanishes in the
same coarse-grid regime. Coarse explicit integration can introduce errors in
both the sampled distribution and its gradients even when every step remains
locally invertible.

\begin{figure*}[t]
\centering
\includegraphics[width=0.70\textwidth]{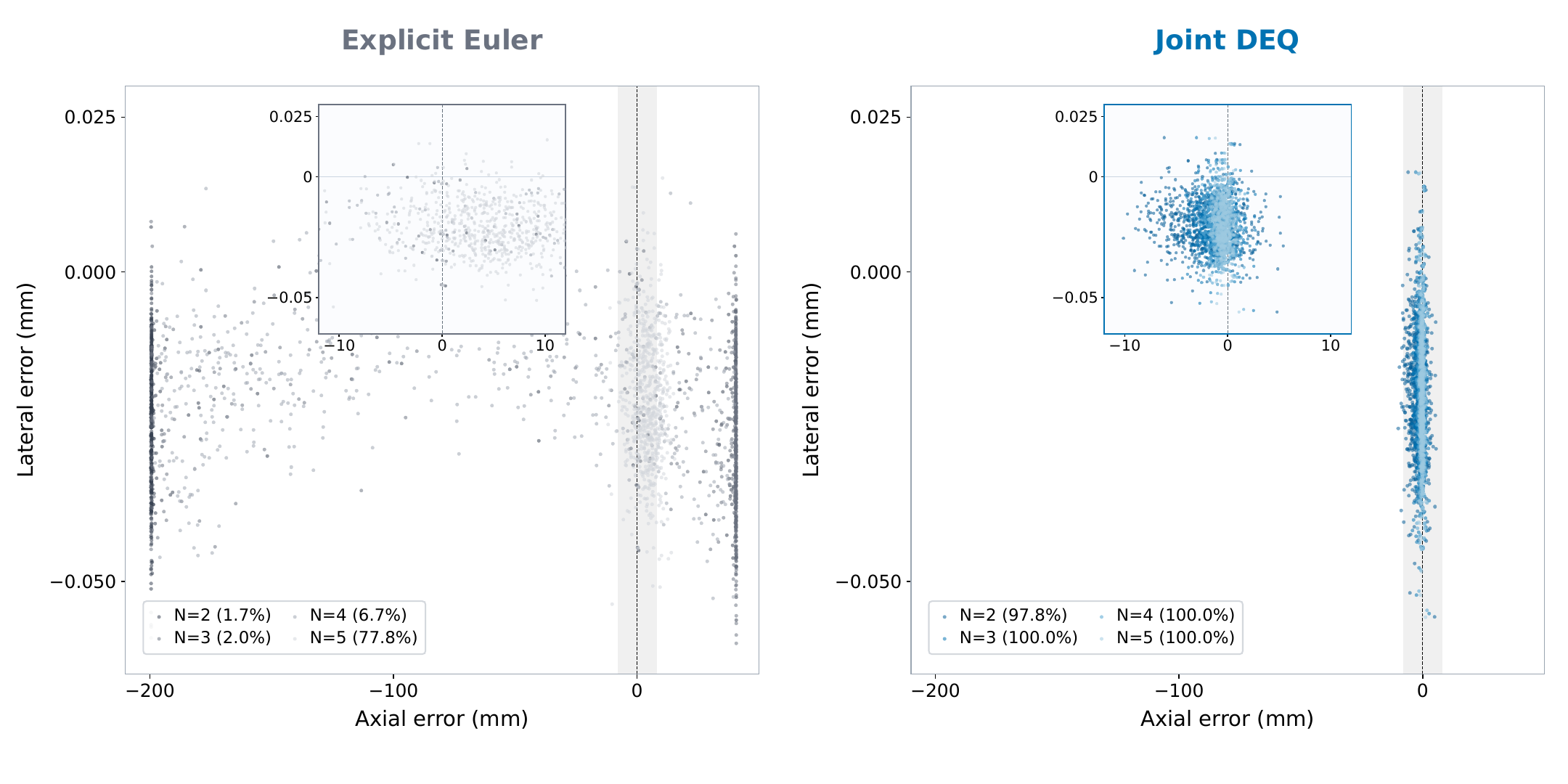}
\caption{Panda reaching errors with explicit Euler (left) and joint implicit
Euler (right), using 2--5 flow steps. Percentages show
success over 640 paired trials per setting: final 3D error below $8$~mm.}
\label{fig:n2-robot}
\end{figure*}

\paragraph{A coarse-step empirical illustration.}
Figure~\ref{fig:n2-robot} shows that implicit Euler achieves more accurate
robot reaching than explicit Euler at low step counts for the tested fixed
contracting flow. With just two steps, the joint DEQ sampler concentrates
endpoints near the target and achieves $97.8\%$ success, versus $1.7\%$ for
explicit Euler. This advantage persists at three to five steps, where implicit
sampling achieves $100\%$ success and explicit Euler reaches at most $77.8\%$.
The comparison uses a fixed field without a trained neural policy.

Together, the analysis and measurements motivate implicit discretization. Its
local sensitivity is governed by $(I-\Delta t\,J)^{-1}$ rather than $I+\Delta t\,J$, which can
preserve concentration when the field contracts toward a target.
The accuracy of implicit steps also depends on the conditioning of the underlying solve.

\section{Method}
\label{sec:method}

We introduce a deep equilibrium flow policy (DEFlow), defined by implicit Euler integration and trained within soft actor--critic (SAC) \citep{haarnoja2018sac,haarnoja2018sacapps}. Following Sec.~\ref{sec:problem}, we sample actions by solving the implicit Euler equations for all integration steps together. We compute policy gradients by implicit differentiation and action log-densities with the change-of-variables formula.

\subsection{Implicit Flow Policy}
\label{sec:field}

For a state $s$, we transform a base sample $a_0=z\sim p_0=\mathcal N(0,\sigma_z^2 I)$ into an action through a state-conditioned velocity field $v_\theta(a,s,t)$. We take $\sigma_z>0$, and the base distribution is independent of $\theta$. We write its density as $p(z)$. We parameterize the velocity field using the gated form of Flow-G in SAC-Flow \citep{zhang2025sacflow},
\begin{equation}
v_\theta(a,s,t) = z_g(a,s,t)\odot\bigl(\tilde h(a,s,t)-a\bigr),
\label{eq:field}
\end{equation}
where $z_g\in(0,1)^{d_a}$ is a sigmoid gate, $\tilde h\in\mathbb R^{d_a}$ is a scaled $\tanh$ output, and $\odot$ denotes elementwise multiplication. Both outputs depend on $(a,s,t)$ and share parameters $\theta$.

Let $\Delta t=1/T$ and $t_k=k\Delta t$. We replace the explicit Euler rollout of SAC-Flow with implicit (backward) Euler \citep{hairer1996solving}:
\begin{equation}
\begin{aligned}
a_k &= a_{k-1}+\Delta t\,v_\theta(a_k,s,t_k),\qquad k=1,\dots,T,\\
a_{\mathrm{env}} &= c\odot\tanh(a_T)+b.
\end{aligned}
\label{eq:implicit-step}
\end{equation}
The field is evaluated at the new point $a_k$. The final squashing transformation follows SAC \citep{haarnoja2018sacapps}, with $c=(a_{\max}-a_{\min})/2>0$ and $b=(a_{\max}+a_{\min})/2$ enforcing the action bounds. Unless stated otherwise, we use $T=4$ integration steps.

Implicit Euler is a standard choice for stiff dynamics \citep{hairer1996solving}. For the scalar linear test equation $\dot x=\lambda x$ with $\operatorname{Re}\lambda<0$, its amplification factor $(1-\Delta t\,\lambda)^{-1}$ has magnitude at most one for every $\Delta t>0$, whereas explicit Euler requires $|1+\Delta t\,\lambda|\leq 1$. The finite-step analysis in Sec.~\ref{sec:problem} motivates our use of implicit Euler. For strongly contracting dynamics, explicit Euler may require very small integration steps to remain stable. In the scalar linear setting above, implicit Euler remains stable for any positive step size.

\subsubsection{Joint Fixed-Point Formulation}
\label{sec:stack}
Summing the first $k$ implicit Euler updates in Eq.~\eqref{eq:implicit-step}, with $a_0=z$, expresses each intermediate action in terms of the base sample and the accumulated velocity:
\begin{equation}
a_k=z+\Delta t\sum_{j=1}^{k}v_\theta(a_j,s,t_j),\qquad k=1,\dots,T.
\label{eq:cumulative-steps}
\end{equation}
We collect the intermediate actions into the vector $Z=(a_1^\top,\dots,a_T^\top)^\top\in\mathbb R^D$, where $D=T d_a$. Define $g_\theta(Z;s,z)$ by stacking the right-hand sides of Eq.~\eqref{eq:cumulative-steps} in the same order. The implicit Euler equations can then be written as a single fixed-point equation:
\begin{equation}
Z^\star=g_\theta(Z^\star;s,z).
\label{eq:stacked-fp}
\end{equation}
The solution $Z^\star$ contains the intermediate actions $a_1^\star,\dots,a_T^\star$ satisfying all implicit Euler steps. Following the DEQ approach \citep{bai2019deq}, we compute this solution with a numerical root finder.

The forward pass searches for a root of $g_\theta(Z;s,z)-Z=0$. We use Anderson acceleration \citep{walker2011anderson}, initialized with every trajectory block equal to $z$, to approximate this root. Each evaluation of $g_\theta$ computes the velocities at the current estimates of the intermediate actions and accumulates them along the trajectory. The final block supplies $a_T$, which is squashed into the environment action by Eq.~\eqref{eq:implicit-step}.

\subsubsection{Backward Pass}
\label{sec:backward}

Following the implicit differentiation procedure of \citet[Eqs.~(8) and~(11)]{bai2019deq}, we compute policy gradients by solving a linear system at the fixed point, without backpropagating through the root-finding iterations. For a differentiable sample loss $\ell(\theta,Z^\star)$, let $A=I-\partial_Z g_\theta$, with derivatives evaluated at $(Z^\star;s,z)$ for fixed $(s,z)$. Assuming a differentiable solution branch and an invertible $A$, we solve for the adjoint vector $\lambda\in\mathbb R^D$ and evaluate the parameter gradient as
\begin{equation}
A^\top\lambda=\nabla_Z\ell,\qquad
\nabla_\theta\ell=
\left.\partial_\theta\ell\right|_{Z^\star}
+(\partial_\theta g_\theta)^\top\lambda.
\label{eq:implicit-backward}
\end{equation}
Gradients are column vectors. The direct derivative $\left.\partial_\theta\ell\right|_{Z^\star}$ holds the trajectory fixed and accounts for the explicit parameter dependence of the log-density and Jacobian penalty introduced below. The adjoint solve avoids forming $A^{-1}$.

In our implementation, we compute $\lambda$ with a batched dense linear solve. Appendix~\ref{app:implicit-differentiation} gives the derivation, the structure of $A$, and full regularity assumptions. Appendix~\ref{app:numerical-implementation} describes the finite-tolerance evaluation and numerical modifications.

\subsection{Maximum-Entropy Policy Optimization}

To train the implicit flow policy with SAC, we need the log-density of its actions \citep{haarnoja2018sacapps}. We derive this density from the discrete implicit Euler steps and use it in the actor--critic updates below.

\subsubsection{Exact Density by Discrete Change of Variables}
\label{sec:density}

To obtain the exact density we use the change-of-variables identity for invertible maps \citep{rezende2015variational,papamakarios2021normalizing}: for $x\mapsto y=f(x)$,
\begin{equation}
p_Y(y) = p_X\bigl(f^{-1}(y)\bigr)\,\bigl|\det \partial f^{-1}/\partial y\bigr|
       = p_X(x)\big/\bigl|\det \partial f/\partial x\bigr|,
\label{eq:cov-general}
\end{equation}
Under the smoothness and invertibility assumptions in Appendix~\ref{app:density-derivation}, each implicit step in Eq.~\eqref{eq:implicit-step} defines a locally invertible map around the selected solution. Differentiating $a_k = a_{k-1} + \Delta t\,v_\theta(a_k,s,t_k)$ gives the differential $da_k = da_{k-1} + \Delta t\,J_k\,da_k$, so $(I-\Delta t\,J_k)\,da_k = da_{k-1}$ and the one-step Jacobian is
\begin{equation}
\frac{\partial a_k}{\partial a_{k-1}} = (I-\Delta t\,J_k)^{-1},\qquad J_k = \left.\frac{\partial v_\theta}{\partial a}\right|_{(a_k,t_k)}.
\label{eq:cov-step}
\end{equation}
Substituting Eq.~\eqref{eq:cov-step} into Eq.~\eqref{eq:cov-general}, a single step changes the log-density by $\log p(a_k) = \log p(a_{k-1}) + \log\bigl|\det(I-\Delta t\,J_k)\bigr|$. The positive sign follows from the inverse in Eq.~\eqref{eq:cov-step}. Composing the $T$ steps into the rollout $z=a_0\to a_1\to\cdots\to a_T$, the log-density is the base density plus one log-determinant per step.
Applying the final squashing transformation gives the environment action $a_{\mathrm{env}} = c \odot \tanh(a_T) + b$ (Eq.~\eqref{eq:implicit-step}). Its exact log-density is therefore:
\begin{equation}
\log \pi_\theta(a_{\mathrm{env}}\mid s) = \log p(z) + \sum_{k=1}^{T} \log\bigl|\det(I - \Delta t\, J_k)\bigr| - \sum_{i=1}^{d_a} \log\bigl[c_i\,(1-\tanh^2 a_{T,i})\bigr]
\label{eq:cov}
\end{equation}
Equation~\eqref{eq:cov} gives the exact action log-density for the selected solution branch under the assumptions above. SAC-Flow uses a joint path log-density over intermediate actions as its entropy term \citep{zhang2025sacflow}, whereas DIME uses a lower bound on the policy entropy \citep{celik2025dime}. Here, the change-of-variables formula gives the log-density of the final action directly. The implementation uses finite solver tolerances, diagonal shifts, and clipping, which can modify the evaluated log-density and policy gradient. The full derivation is given in Appendix~\ref{app:density-derivation}.

\subsubsection{Actor--Critic Updates}
\label{sec:recalibration}

We train DEFlow within the SAC framework \citep{haarnoja2018sacapps}, using the action samples and log-densities derived above. Let $a_\theta(s,z)=c\odot\tanh(a_T^\star)+b$ denote an action sampled from the implicit policy. We compute the soft Bellman target as
\begin{equation}
y=r+\gamma\left[
\min_{j=1,2}Q_{\bar\phi_j}(s',a')-\alpha\log\pi_\theta(a'\mid s')
\right],
\label{eq:critic-target}
\end{equation}
where $a'=a_\theta(s',z')$, $z'\sim p_0$, and $d$ is the termination indicator. Here $\bar\phi_j$ denotes the parameters of target critic $j$, and the action log-density is given by Eq.~\eqref{eq:cov}. For terminal transitions, we set $y=r$. Each critic is trained by minimizing the mean squared error
\begin{equation}
\mathcal L_Q(\phi_j)=
\mathbb E_{(s,a,r,s',d)\sim\mathcal D,\,z'\sim p_0}
\left[\bigl(Q_{\phi_j}(s,a)-y\bigr)^2\right],
\qquad j=1,2.
\label{eq:critic-loss}
\end{equation}

We update the implicit flow policy by minimizing the SAC actor objective \citep{haarnoja2018sac,haarnoja2018sacapps}:
\begin{equation}
\mathcal L_\pi(\theta)=
\mathbb E_{s\sim\mathcal D,\,z\sim p_0}
\left[\alpha\log\pi_\theta(a_\theta(s,z)\mid s)
-\min_{j=1,2}Q_{\phi_j}(s,a_\theta(s,z))\right].
\label{eq:implicit-actor-loss}
\end{equation}
The log-density is evaluated by Eq.~\eqref{eq:cov}. We compute the policy gradient using the implicit backward pass in Eq.~\eqref{eq:implicit-backward}. In the regularized configuration, we minimize $\mathcal L_\pi+\beta\mathcal R_J$, where $\beta\geq0$ weights a per-step Jacobian penalty motivated by Jacobian regularization in DEQ models \citep{bai2021jacobian}. Appendix~\ref{app:jacobian-regularization} defines $\mathcal R_J$ and specifies its hyperparameters.

The target critics are initialized with $\bar\phi_j=\phi_j$ and updated by
exponential moving averages $\bar\phi_j\leftarrow(1-\tau)\bar\phi_j+\tau\phi_j$
for $j=1,2$, where $\tau\in(0,1]$ is the target averaging rate
\citep{haarnoja2018sacapps}.

\subsubsection{Algorithm}
\label{sec:algorithm}

Algorithm~\ref{alg:deflow} in Appendix~\ref{app:implementation} summarizes data collection and actor--critic updates following SAC \citep{haarnoja2018sacapps}. Each action $a_\theta(s,z)$ is obtained by solving Eq.~\eqref{eq:stacked-fp} with Anderson acceleration and applying the squashing transformation in Eq.~\eqref{eq:implicit-step}. Policy updates use the log-density in Eq.~\eqref{eq:cov} and the implicit gradients in Eq.~\eqref{eq:implicit-backward}. The pseudocode takes the current temperature $\alpha$ and regularization weight $\beta$ as inputs to each update. Numerical evaluation is described in Appendix~\ref{app:numerical-implementation}.

\section{Experiments}
\label{sec:experiments}

\subsection{Experimental Setup}
\label{sec:setup}

We evaluate DEFlow in three settings: four variants of MultiGoal-4 with
different navigation and docking tolerances \citep{haarnoja2017sql}, five MuJoCo locomotion tasks
\citep{todorov2012mujoco}, and five manipulation tasks: PickCube and PushCube
in ManiSkill \citep{tao2025maniskill3}, and ButtonPressWall, PegInsertSide, and
PushWall in MetaWorld \citep{yu2020metaworld}. Training budgets are 200k
transitions for MultiGoal, 1M environment steps for MuJoCo, and 500k recorded
steps for manipulation. Our main comparison methods are SAC \citep{haarnoja2018sac},
SAC-Flow, DIME, and QSM. For ablation studies, we compare DEFlow with a matched
explicit-Euler flow on MultiGoal and vary the number of integration steps $T$.
All baseline
results come from our own implementations and training runs. Evaluation protocols,
task-specific configurations, and additional
statistical analyses are provided in Appendices~\ref{app:implementation}
and~\ref{app:results}.

\subsection{Benchmark Results}
\label{sec:matched}

\paragraph{Successful diversity under precision constraints.}
\label{sec:multigoal}
We extend the four-goal MultiGoal task \citep{haarnoja2017sql} to test whether an expressive policy can combine diverse goal choices with precise actions. The agent receives a dense distance reward while navigating toward one of four symmetric goals. On entering a goal region, it performs one docking action and receives a bonus if the action is sufficiently close to the goal's target. We study four variants combining wide or narrow goal regions with loose or tight docking tolerances. Wide goal regions are circles of radius 1, while narrow regions are radial ellipses with semi-axes 1 and 0.20. Docking succeeds when the final action lies within distance 0.12 (loose) or 0.06 (tight) of the target associated with the reached goal. This design varies the precision required during navigation and during the final docking action separately. Figure~\ref{fig:multigoal} in Appendix~\ref{app:tables} illustrates the four variants. To maintain several successful strategies, the policy must choose different goals across episodes and concentrate its final action around the target associated with each goal.

Our main metric, successful coverage (U4), measures how many distinct goals a policy successfully completes within a small number of attempts. Let $p_i$ be the probability that an episode from the origin reaches goal $i$ and docks successfully. For four independent episodes under the same policy,
\begin{equation}
\mathrm{U4}=\frac{1}{4}\sum_{i=1}^{4}\left[1-(1-p_i)^4\right].
\label{eq:docking-u4}
\end{equation}
Here, $1-(1-p_i)^4$ is the probability of completing goal $i$ at least once, so U4 is the expected fraction of distinct goals successfully completed. It rewards both reliable docking and coverage of alternative goals. Always succeeding at one goal gives U4 $=0.25$, whereas uniform success across all four goals attains the maximum $1-(3/4)^4\approx0.684$. We estimate $p_i$ from 1,000 stochastic episodes per trained policy and average U4 across training seeds.

DEFlow achieves the highest mean successful coverage on all evaluated variants (Table~\ref{tab:docking}). On loose docking, it approaches the balanced four-goal ceiling. Its mean successful coverage exceeds that of the second-best method by up to $17\%$. Thus, DEFlow retains the strongest combination of diversity and successful completion when the final action must be more precise. Figure~\ref{fig:docking-learning} also shows that DEFlow develops successful coverage early in training. Appendix~\ref{app:docking-failures} and Sec.~\ref{sec:ledger} examine the failures of explicit flow and QSM.

\begin{center}
\begin{minipage}{\textwidth}
\centering
\small
\captionof{table}{Successful coverage (U4) at 200k transitions: mean $\pm$ SD over completed seeds, 1,000 episodes per seed. Bold/underline mark the top two means.}
\label{tab:docking}
{\scriptsize\setlength{\tabcolsep}{3pt}
\begin{tabular}{@{}llrrrrrr@{}}
\toprule
Navigation & Docking & SAC & SAC-Flow & DIME & QSM & Explicit flow & DEFlow \\
\midrule
Wide & Loose & $0.397\pm0.127$ & $0.522\pm0.224$ & $\underline{0.634}\pm0.021$ & $0.000\pm0.000$ & $0.507\pm0.030$ & $\mathbf{0.671}\pm0.002$ \\
Narrow & Loose & $0.376\pm0.019$ & $0.613\pm0.032$ & $\underline{0.632}\pm0.012$ & $0.000\pm0.000$ & $0.499\pm0.100$ & $\mathbf{0.661}\pm0.020$ \\
Wide & Tight & $0.428\pm0.105$ & $0.481\pm0.067$ & $\underline{0.502}\pm0.011$ & $0.000\pm0.000$ & $0.412\pm0.084$ & $\mathbf{0.589}\pm0.128$ \\
Narrow & Tight & $0.467\pm0.074$ & $0.451\pm0.084$ & $\underline{0.491}\pm0.009$ & $0.000\pm0.000$ & $0.417\pm0.153$ & $\mathbf{0.552}\pm0.113$ \\
\bottomrule
\end{tabular}
}
\end{minipage}
\end{center}

\begin{center}
\begin{minipage}{\textwidth}
\centering
\includegraphics[width=\textwidth]{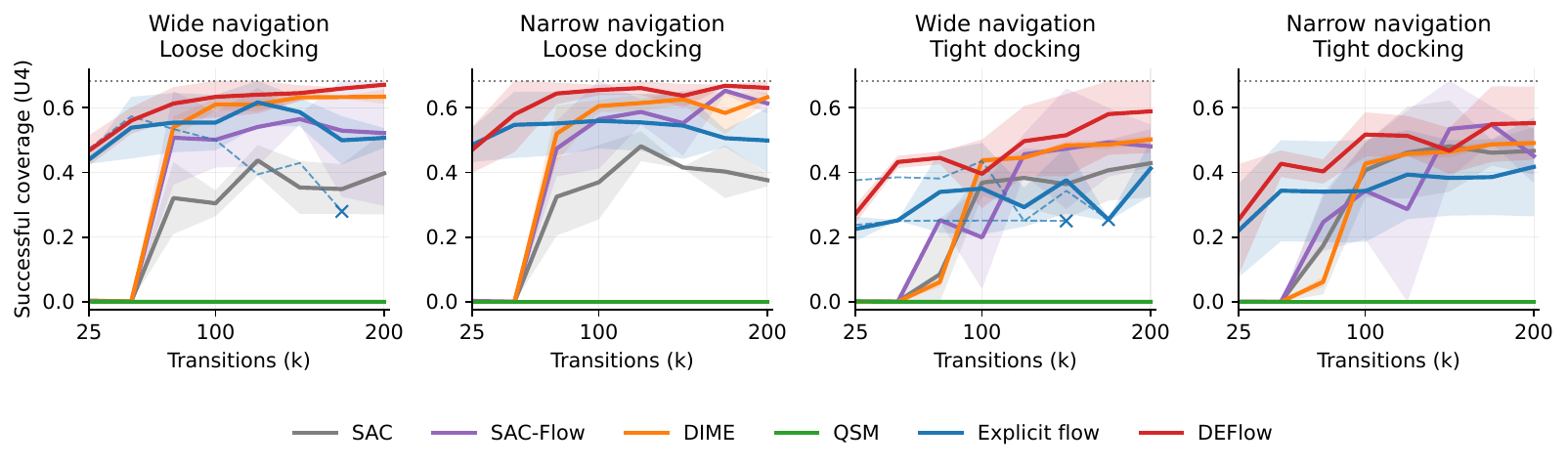}
\captionof{figure}{Successful coverage during training (mean $\pm$ SD). The dotted line marks the maximum U4.}
\label{fig:docking-learning}
\end{minipage}
\end{center}

\paragraph{Continuous control.}
Figure~\ref{fig:main} shows learning curves across five MuJoCo tasks, and Table~\ref{tab:mean-return} in Appendix~\ref{app:tables} summarizes return at the recorded 1M evaluation. We report means and standard deviations across three runs per task. Appendix~\ref{app:tables} provides per-seed results, confidence intervals, and sensitivity analyses at alternative evaluation horizons. These tasks do not explicitly require multimodal action selection, and we observe no consistent advantage from greater policy expressivity, with SAC, SAC-Flow, DIME, and DEFlow achieving broadly comparable returns overall and SAC retaining a small lead in the sum of returns across tasks.

\begin{center}
\begin{minipage}{\textwidth}
\centering
\includegraphics[width=\textwidth]{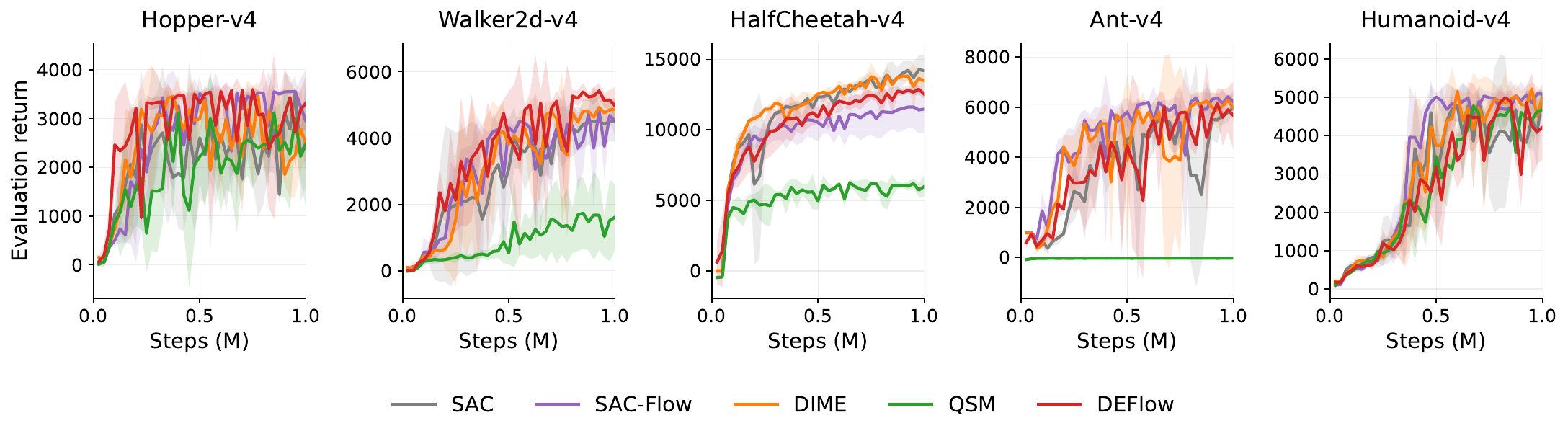}
\captionof{figure}{MuJoCo learning curves. Lines show run means, bands show $\pm1$ sample SD.}
\label{fig:main}
\end{minipage}
\end{center}

\paragraph{Manipulation.}
Manipulation tasks test how quickly a policy turns contact with an object into useful behavior. We compare SAC, SAC-Flow, DIME, and DEFlow on five tasks from ManiSkill and MetaWorld, with a budget of 500k recorded training steps. Figure~\ref{fig:manip} shows deterministic evaluation returns, and Table~\ref{tab:manip-return} in Appendix~\ref{app:tables} summarizes their time average over training. Table~\ref{tab:manip} in Appendix~\ref{app:tables} reports final sampled ever-success: the fraction of evaluation episodes in which the policy completes the task at least once.

DEFlow's main advantage is early learning under the evaluated training recipes. It achieves the highest mean time-averaged return on four of the five tasks. On PickCube, the advantage also appears in task completion: DEFlow reaches higher sampled success at the intermediate checkpoints, before DIME catches up later in training. On PushCube, DEFlow attains high return by the first evaluation and maintains it through the rest of training. The return curves and early success evaluations thus capture gains that a final-checkpoint comparison alone would miss. At the final checkpoint, DIME has the highest sampled-success means on the cube tasks, while SAC leads on PegInsertSide.

\begin{center}
\begin{minipage}{\textwidth}
\centering
\includegraphics[width=\textwidth]{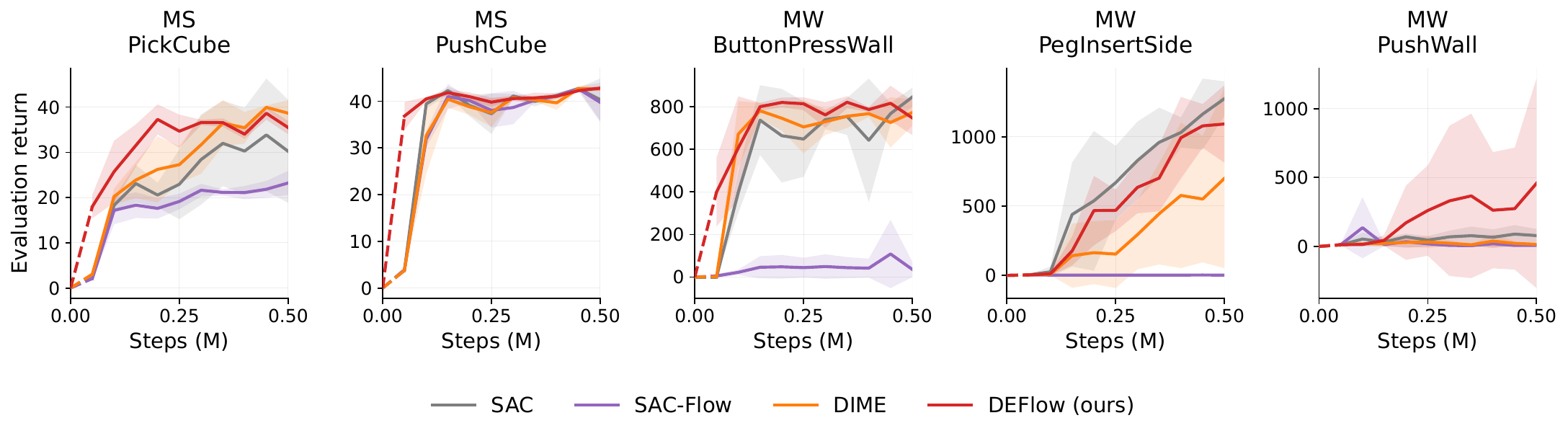}
\captionof{figure}{Manipulation evaluation returns, mean $\pm$ SD across three training seeds. Dashed segments connect an illustrative origin to the first evaluation.}
\label{fig:manip}
\end{minipage}
\end{center}

\subsection{Ablation Studies}
\label{sec:ledger}

\paragraph{Implicit vs. explicit flow.}
\label{sec:implicit-explicit}
To test the role of implicit integration during policy learning, we replace DEFlow's implicit-Euler updates with explicit Euler while retaining the actor architecture, four integration steps, discrete density objective, and training settings. Both policies are trained independently on the docking tasks. DEFlow achieves higher successful coverage on loose docking (Table~\ref{tab:docking}), and tight docking exposes a further limitation of the explicit variant: it either concentrates on a preferred goal or fails numerically (details in Appendix~\ref{app:docking-failures}). This comparison strongly supports implicit integration for learning diverse successful behavior under a narrow action tolerance.

\paragraph{Integrator step-count ablation.}
\label{sec:kablation}
On the MultiGoal variant with wide goal regions and tight docking, we freeze two policies trained with $T=4$ and change only the number of docking integration steps. Each policy keeps its own integrator and uses the same states and Gaussian inputs across step counts. At the training depth $T=4$, the explicit policy achieves only $50.1\%$ balanced docking accuracy on a shared bank of reachable states, with near-zero accuracy for two of the four goals, while DEFlow achieves $94.1\%$. Even on each policy's own visited states, changing only the docking integration depth from $T=4$ to $T=2$ reduces explicit success from $99.8\%$ to $0.7\%$, versus $96.5\%$ to $53.7\%$ for DEFlow, and increasing $T$ also degrades the explicit policy. For this pair, DEFlow is more robust to changes in step count (Figures~\ref{fig:kabl-shared} and~\ref{fig:kabl-own}).

\section{Limitations and Costs}
\label{sec:drawbacks}

\paragraph{Training time.}
In our measured runs, DEFlow had roughly $40\%$ lower effective training throughput than SAC-Flow and took about three times as long as SAC. Its wall-clock cost was of the same order as DIME's. These comparisons are approximate and depend on workload and shared hardware. Each implicit actor update requires a nonlinear forward solve and a dense linear backward solve.

\paragraph{Exactness is tolerance-limited in training.}
Equation~\eqref{eq:cov} is exact for a solution of the implicit equations under the assumptions in Appendix~\ref{app:density-derivation}. The implementation uses finite solver tolerances and numerical modifications (Appendix~\ref{app:numerical-implementation}). The trained-policy audit in Appendix~\ref{sec:verify} measures the resulting discrepancies at final manipulation checkpoints. The audit finds numerical inaccuracy, especially in PushWall actor gradients. Tightening the solver at evaluation reduces these errors, while sampled task success changes little. We use the original solver settings for benchmark evaluations to avoid the extra cost of tighter solves.

\section{Conclusion}
We introduced DEFlow, a deep-equilibrium formulation of flow policies for maximum-entropy reinforcement learning. By combining implicit sampling, tractable density evaluation, and implicit differentiation, DEFlow provides an alternative to explicit flow actors. Our experiments highlight its potential for expressive control and the remaining trade-offs in numerical accuracy and computational cost.

\bibliographystyle{iclr2027_conference}
\bibliography{references}

\appendix
\section{Additional Derivations}
\label{app:derivations}

\subsection{Finite-Step Analysis}
\label{app:finite-step}

This appendix records the assumptions and proof details behind the claims in
Sec.~\ref{sec:problem}. We use $T\in\mathbb N_+$ integration steps on $[0,1]$ with $\Delta t=1/T$
and $t_k=k\Delta t$. We fix a state and suppress it from the notation.
The actions are unsquashed, the base variable is shared when two rollouts are
compared, and the field is the ideal independent-interpolation field. These
results isolate the numerical effects of finite-step integration on
analytically specified vector fields.

\subsubsection{A smooth multimodal target can require a fine grid}
\label{app:mixture-proof}

Consider
\begin{equation}
 A=mC+\sigma G,\qquad C\in\{-1,1\},\qquad
 I_t=(1-t)Z+tA,
 \label{eq:app-mixture}
\end{equation}
where $m,\sigma>0$, $\mathbb P(C=\pm1)=1/2$, and $Z,G$ are standard Gaussians.
The variables $C,Z,G$ are mutually independent. Small-width limits keep $m$ fixed. The
terminal target is
$\pi^\star_{m,\sigma}=\tfrac12\mathcal N(-m,\sigma^2)+\tfrac12\mathcal N(m,\sigma^2)$.
Writing $q_t=(1-t)^2+\sigma^2t^2$, conditional Gaussian regression gives
\begin{equation}
 v^\star(x,t)=b_tx+d_t\tanh(k_tx),\quad
 b_t=\frac{\sigma^2t-(1-t)}{q_t},\quad
 d_t=\frac{m(1-t)}{q_t},\quad k_t=\frac{mt}{q_t}.
 \label{eq:app-exact-field}
\end{equation}
The denominator is bounded away from zero, so this field is globally well
posed. Its derivative is
\begin{equation}
 \partial_xv^\star(x,t)=b_t+d_tk_t\operatorname{sech}^2(k_tx).
 \label{eq:app-field-derivative}
\end{equation}
For $0<\sigma\le\min\{1/2,m/10\}$, evaluating at $x=0$ and
$t=(1+\sigma)^{-1}$ yields
\begin{equation}
 \sup_{x,t}|\partial_xv^\star(x,t)|
 \ge \frac{m^2}{8\sigma^3}.
 \label{eq:app-lipschitz-lower}
\end{equation}
Thus the exact field remains regular for every fixed $\sigma>0$, while the
near-identity requirement for a discrete step becomes increasingly demanding
as the target narrows.

For completeness, let $J_k=D_av(a_k,t_k)$ and $B_k=I+\Delta t\,J_k$. If
$\|J_k\|\le L_k$, the triangle and reverse-triangle inequalities give
\begin{equation}
 [1-\Delta t\,L_k]_+\le\sigma_{\min}(B_k)\le\sigma_{\max}(B_k)
 \le1+\Delta t\,L_k.
 \label{eq:app-step-singular-values}
\end{equation}
When $\Delta t\,L_k<1$ for all $k$, multiplication of singular-value bounds gives
\begin{equation}
 \kappa_2(B_{T-1}\cdots B_0)
 \le\prod_{k=0}^{T-1}\frac{1+\Delta t\,L_k}{1-\Delta t\,L_k}.
 \label{eq:app-rollout-conditioning}
\end{equation}
The same calculation for an implicit step produces the inverse factor
$(I-\Delta t\,J_{k+1})^{-1}$, which is why the two discretizations can have different
conditioning on the same field.

\paragraph{Sampling error.}
For the explicit Euler discretization of~\eqref{eq:app-exact-field}, define
$\pi_{\Delta t}^\star$ as the law after $T=1/\Delta t$ steps. The linear part of the update
contracts the outward fluctuation. An induction gives
\begin{equation}
 |\widehat X_k|\le mt_k+R_k|Z|,
 \qquad R_{k+1}=\left(1+\Delta t\,b_{t_k}\right)R_k,\quad R_0=1.
 \label{eq:app-envelope}
\end{equation}
Convexity of $\sqrt{q_t}$ implies
\begin{equation}
 R_T\le
 \frac{\sigma^2(1-\Delta t)}{\sqrt{(\Delta t)^2+\sigma^2(1-\Delta t)^2}}
 \le\frac{\sigma^2}{\Delta t}.
 \label{eq:app-tail-factor}
\end{equation}
Now use the 1-Lipschitz function $f_m(x)=(|x|-m)_+$. The envelope gives
$\|f_m(\widehat X_T)\|_{L^2}\le R_T$, while the target satisfies
$\|f_m(A)\|_{L^2}\ge\sigma/\sqrt2$. For any coupling of the numerical and
target variables, the reverse triangle inequality in $L^2$ therefore gives
\begin{equation}
 W_2(\pi_{\Delta t}^\star,\pi^\star_{m,\sigma})
 \ge\left[\frac{\sigma}{\sqrt2}-R_T\right]_+.
 \label{eq:app-wasserstein-general}
\end{equation}
Substitution of~\eqref{eq:app-tail-factor} proves
\begin{equation}
 \Delta t\ge2\sqrt2\,\sigma
 \quad\Longrightarrow\quad
 W_2(\pi_{\Delta t}^\star,\pi^\star_{m,\sigma})
 \ge\frac{\sigma}{2\sqrt2}.
 \label{eq:app-wasserstein-bound}
\end{equation}
At one step, $v^\star(x,0)=-x$, so $\widehat X_1=0$ and
$W_2(\pi_1^\star,\pi^\star_{m,\sigma})=\sqrt{m^2+\sigma^2}$. At two steps,
direct substitution gives
\begin{equation}
 \widehat X_2=
 \frac{\sigma^2Z+m\tanh\!\left(mZ/(1+\sigma^2)\right)}{1+\sigma^2}.
 \label{eq:app-two-step-map}
\end{equation}
As $\sigma\downarrow0$, this converges in $L^2$ to
$m\tanh(mZ)$, whereas the target converges to the two-point law on
$\{-m,m\}$. The limiting Wasserstein error is consequently
\begin{equation}
 m\left(\mathbb E\left[(1-\tanh(m|Z|))^2\right]\right)^{1/2}>0.
 \label{eq:app-two-step-error}
\end{equation}
This establishes intermodal shape distortion without invoking a singular
explicit update.

\subsubsection{A fixed budget can distort the policy gradient}
\label{app:gradient-proof}

The gradient statement in Sec.~\ref{sec:problem} uses the centered Gaussian
subfamily $A\sim\mathcal N(0,\vartheta)$, with $\sigma=\sqrt\vartheta$, and
the terminal loss $\ell(a)=a^2/2$. The exact flow gives
$X_1=\sqrt\vartheta Z$ and $\mathcal L'(\vartheta)=1/2$. For $T\ge2$, direct
substitution of $t_k=k/T$ into the explicit update yields
\begin{equation}
 R_T(\vartheta)=\prod_{k=0}^{T-1}
 \frac{(T-k)(T-k-1)+\vartheta k(k+1)}{(T-k)^2+\vartheta k^2},
 \qquad \widehat X_T=R_T(\vartheta)Z.
 \label{eq:app-gradient-product}
\end{equation}
Hence $\mathcal L_{\Delta t}'(\vartheta)=R_TR_T'$. Setting
$u=T^2\vartheta=(\sigma/\Delta t)^2$, the factors are nondecreasing and their
logarithmic derivatives give
\begin{equation}
 0\le 2R_TR_T'\le2u e^{2u}(1+u)\le3u,
 \qquad 0\le u\le\frac18.
 \label{eq:app-gradient-bound}
\end{equation}
Since the exact derivative is $1/2$, this is the claimed bound on the
relative gradient. The statement is deliberately limited to this terminal
loss and parameterization; it identifies a possible rollout-sensitivity
mechanism rather than characterizing the full SAC actor gradient.

\subsubsection{Why architectural stabilization alone is insufficient}
\label{app:stabilization-counterexamples}

Two short counterexamples clarify the scope of the preceding results. First,
clipping an optimizer step bounds $\|\theta^{r+1}-\theta^r\|$, but does not
bound $\|D_av_\theta\|$, the smallest singular value of a step map, or the
sampling error. For $v_\lambda(x)=-\lambda x$, explicit Euler gives
\begin{equation}
 X_1=e^{-\lambda}Z,\qquad
 \widehat X_T=(1-\Delta t\lambda)^TZ,\qquad
 W_2=\left||1-\Delta t\lambda|^T-e^{-\lambda}\right|.
 \label{eq:app-linear-counterexample}
\end{equation}
The continuous flow is well posed, while $\lambda=1/\Delta t$ makes one step
singular and $\lambda>2/\Delta t$ amplifies its sensitivity. A clipped update can be
zero at a parameter-space minimizer, so clipping does not exclude this field.

Second, a bounded gated field can keep every Euler state in a bounded range
while its gate derivative makes the step Jacobian singular or arbitrarily
ill-conditioned. Normalized features and a bounded output head do not remove
this possibility. These examples show why the method must expose and control
the discrete step structure; they do not imply that the trained actor always
enters the counterexample regime.

\subsection{Implicit Differentiation}
\label{app:implicit-differentiation}
\label{app:implicit-assumptions}

We derive the backward pass in Sec.~\ref{sec:backward} following \citet{bai2019deq}. Fix $(s,z)$ and let $Z^\star$ solve Eq.~\eqref{eq:stacked-fp}. Assume that the velocity field has continuous first and second derivatives in $(a,\theta)$ and that each matrix $I-\Delta t J_k$ is invertible at this solution, where $J_k=\partial_a v_\theta(a_k^\star,s,t_k)$.

Differentiating the fixed-point equation gives
\begin{equation}
A\,\frac{dZ^\star}{d\theta}=\partial_\theta g_\theta,
\qquad A=I-\partial_Z g_\theta.
\label{eq:trajectory-derivative}
\end{equation}
All derivatives are evaluated at $(Z^\star;s,z)$. From Eq.~\eqref{eq:cumulative-steps}, block $k$ depends only on actions $1,\ldots,k$. The blocks of $A$ are therefore
\begin{equation}
A_{kj}=
\begin{cases}
-\Delta t J_j, & j<k,\\
I-\Delta t J_k, & j=k,\\
0, & j>k.
\end{cases}
\qquad
\det A=\prod_{k=1}^{T}\det(I-\Delta t J_k).
\label{eq:det-factorization}
\end{equation}
Thus $A$ is block lower triangular and invertible under the stated assumptions. The implicit function theorem then gives a locally differentiable solution as $(z,\theta)$ varies.

For a differentiable sample loss $\ell(\theta,Z^\star)$, the chain rule gives
\begin{equation}
\nabla_\theta\ell
=\left.\partial_\theta\ell\right|_{Z^\star}
+\left(\frac{dZ^\star}{d\theta}\right)^\top\nabla_Z\ell
=\left.\partial_\theta\ell\right|_{Z^\star}
+(\partial_\theta g_\theta)^\top A^{-\top}\nabla_Z\ell.
\end{equation}
Solving $A^\top\lambda=\nabla_Z\ell$ recovers Eq.~\eqref{eq:implicit-backward}. This avoids forming either $A^{-1}$ or the full derivative $dZ^\star/d\theta$. The direct term $\left.\partial_\theta\ell\right|_{Z^\star}$ includes the dependence of the log-density and Jacobian penalty on $\theta$ with the actions held fixed. Their derivatives involve second derivatives of the velocity field.

To apply this calculation to the expected actor loss, we assume that, for almost every $(s,z)$, the selected solution remains differentiable throughout a common neighborhood of the current parameters. We also assume that the sample-loss derivatives in this neighborhood are bounded by a function of $(s,z)$ with finite expectation. Holding the replay distribution fixed, these conditions allow differentiation under the expectation and justify the minibatch gradient in Algorithm~\ref{alg:deflow}. Numerical evaluation of this gradient is described in Appendix~\ref{app:numerical-implementation}.

\subsection{Action Log-Density}
\label{app:density-derivation}

We justify the change-of-variables calculation in Sec.~\ref{sec:density}. Fix a state $s$. Assume that a measurable, deterministic rule selects an implicit trajectory for almost every $z\sim p_0$. On open regions covering these samples up to a set of probability zero, assume that the selected trajectory is continuously differentiable in $z$ and that every $I-\Delta t J_k$ is invertible.

The log form of the change-of-variables identity in Eq.~\eqref{eq:cov-general} is
\begin{equation}
\log p_Y(y) = \log p_X(x) - \log\bigl|\det \partial f/\partial x\bigr|.
\label{eq:cov-log}
\end{equation}

An implicit Euler step has the explicit inverse relation
\begin{equation}
a_{k-1}=H_k(a_k),
\qquad H_k(a)=a-\Delta t\,v_\theta(a,s,t_k).
\label{eq:app-step-inverse}
\end{equation}
Consequently, the endpoint determines every preceding action and the base sample:
$z=H_1\circ\cdots\circ H_T(a_T)$.
Two different base samples cannot produce the same selected endpoint. The selected map $z\mapsto a_T^\star$ is therefore one-to-one, even if the step equations admit other solutions. The pre-squash density is defined on the image of this selected map and is zero outside it.

Since $\partial_a H_k=I-\Delta t J_k$, differentiating the inverse map gives
\begin{equation}
\frac{\partial z}{\partial a_T}
=(I-\Delta t J_1)\cdots(I-\Delta t J_T).
\end{equation}
Composing the step maps and applying the log form of change of variables gives
\begin{equation}
\frac{\partial a_T}{\partial z} = (I - \Delta t\, J_T)^{-1}\cdots(I - \Delta t\, J_1)^{-1},
\qquad
\log \pi(a_T\mid s) = \log p(z) + \sum_{k=1}^{T} \log\bigl|\det(I - \Delta t\, J_k)\bigr|.
\label{eq:cov-jac}
\end{equation}
Here, $\pi(a_T\mid s)$ denotes the density before squashing. The final transformation
$a_{\mathrm{env}}=c\odot\tanh(a_T^\star)+b$ is one-to-one for $c_i>0$, with Jacobian determinant
$\prod_i c_i(1-\tanh^2(a_{T,i}^\star))$.
Dividing by this determinant and taking logarithms gives the final-action log-density:
\begin{equation*}
\log \pi_\theta(a_{\mathrm{env}}\mid s) = \log p(z) + \sum_{k=1}^{T} \log\bigl|\det(I - \Delta t\, J_k)\bigr| - \sum_{i=1}^{d_a} \log\bigl[c_i\,(1-\tanh^2 a_{T,i})\bigr]
\end{equation*}
This is Eq.~\eqref{eq:cov}, including the action-bound correction used in SAC \citep{haarnoja2018sacapps}. This identity concerns an exact solution of the implicit equations. Appendix~\ref{app:numerical-implementation} specifies how it is evaluated at finite solver tolerance.

\clearpage
\section{Implementation Details}
\label{app:implementation}

\subsection{Architecture}
\label{app:architecture}

DEFlow uses the gated velocity field of SAC-Flow \citep{zhang2025sacflow}. Its inputs are the observation, intermediate action, and time embedding. The time embedding maps 32 sinusoidal features through two layers of widths 64 and 32. A shared projection with SiLU and layer normalization feeds separate gate and candidate heads. Each head has one hidden layer and outputs an action-dimensional vector. The gate is sigmoid, and the candidate is a scaled $\tanh$ as in Eq.~\eqref{eq:field}. The velocity width is 256, the candidate scale is 4, and the gate output bias is initialized to 1. Actions start from standard Gaussian noise and use four implicit-Euler steps before the final $\tanh$ and action-bound scaling.

\paragraph{Jacobian regularization.}
\label{app:jacobian-regularization}
Motivated by Jacobian regularization for DEQ models \citep{bai2021jacobian}, we penalize the individual velocity Jacobians:
\begin{equation}
\mathcal R_J(\theta)
=\mathbb E_{s\sim\mathcal D,\,z\sim p_0}
\left[\sum_{k=1}^{T}
\bigl[\|\Delta t J_k\|_2-\sigma^\ast\bigr]_+^2\right],
\qquad [x]_+=\max(x,0),
\label{eq:jacobian-penalty}
\end{equation}
and minimize $\mathcal L_\pi+\beta\mathcal R_J$. We use $\sigma^\ast=0.9$ and increase $\beta$ from zero to $0.01$ over the first $100{,}000$ environment steps.

\paragraph{Temperature updates.}
In the regularized DEFlow recipe, we update $\log\alpha$ from the actor log-densities. The target is an exponential moving average of negative log-density (weight $0.01$) minus $0.7$, with floor $-0.7d_a$. The update winsorizes log-densities at the minibatch 5th and 95th percentiles, leaks $\log\alpha$ toward $\log(0.05)$ with weight $0.01$, and enforces $\alpha\geq0.02$. We also tested fixed-temperature variants, which sometimes reached comparable return but did not provide stable adaptive temperature behavior.

\subsection{Implementation Details}
\label{app:training}

Algorithm~\ref{alg:deflow} summarizes the SAC updates with DEFlow's implicit actor. Table~\ref{tab:implementation} gives one reference configuration for the matched-critic MuJoCo comparison. All methods use twin $4\times512$ ReLU critics in this comparison. We retain each baseline's policy class and actor objective: SAC has a squashed Gaussian density \citep{haarnoja2018sacapps}, SAC-Flow uses its noisy explicit rollout and joint-path entropy term \citep{zhang2025sacflow}, DIME uses its diffusion entropy bound \citep{celik2025dime}, and QSM matches its score to the critic action gradient without a policy log-density \citep{psenka2024qscore}. We implement each baseline's actor and objective in a common training pipeline. Matching critic capacity is necessary to assess actor performance without confounding it with differences in the critic.

\begin{algorithm}[ht]
\caption{DEFlow}
\label{alg:deflow}
\begin{algorithmic}[1]
\REQUIRE Initial policy parameters $\theta$, critic parameters $\phi_1,\phi_2$, minibatch size $B$, target averaging rate $\tau$.
\STATE Initialize target parameters $\bar\phi_j\gets\phi_j$ for $j=1,2$ and replay buffer $\mathcal D\gets\varnothing$.
\STATE Reset the environment and observe the initial state $s$.
\FOR{each iteration}
\FOR{each environment step}
\STATE Draw $z\sim p_0$ and compute $a=a_\theta(s,z)$.
\STATE Execute $a$ and observe reward $r$, next state $s'$, and termination indicator $d$.
\STATE Store $(s,a,r,s',d)$ in $\mathcal D$.
\STATE Set $s\gets s'$, or reset the environment if the episode has ended.
\ENDFOR
\FOR{each gradient step}
\STATE Draw a minibatch $\{(s_i,a_i,r_i,s_i',d_i)\}_{i=1}^{B}$ from $\mathcal D$.
\STATE Draw $z_i'\sim p_0$ and compute $a_i'=a_\theta(s_i',z_i')$ for each $i$.
\STATE Evaluate $\log\pi_\theta(a_i'\mid s_i')$ using Eq.~\eqref{eq:cov}.
\STATE Compute targets $y_i$ using Eq.~\eqref{eq:critic-target}, with $y_i=r_i$ when $d_i=1$.
\STATE Update $\phi_1,\phi_2$ by minimizing the minibatch critic loss in Eq.~\eqref{eq:critic-loss}, holding $y_i$ fixed.
\IF{a policy update is due}
\STATE Draw fresh $z_i\sim p_0$ and compute $\tilde a_i=a_\theta(s_i,z_i)$ for each $i$.
\STATE Evaluate $\log\pi_\theta(\tilde a_i\mid s_i)$ using Eq.~\eqref{eq:cov}.
\STATE Form each sample loss $\ell_i$ using Eq.~\eqref{eq:implicit-actor-loss} and the $\beta$-weighted penalty in Eq.~\eqref{eq:jacobian-penalty}.
\STATE Solve the adjoint systems in Eq.~\eqref{eq:implicit-backward} to compute $\nabla_\theta\ell_i$.
\STATE Update $\theta$ using $B^{-1}\sum_{i=1}^{B}\nabla_\theta\ell_i$.
\STATE Update $\alpha$.
\ENDIF
\STATE Update target parameters $\bar\phi_j\gets(1-\tau)\bar\phi_j+\tau\phi_j$ for $j=1,2$.
\ENDFOR
\ENDFOR
\end{algorithmic}
\end{algorithm}

\begin{table}[ht]
\centering\small
\caption{Reference matched-critic MuJoCo configuration. Method rows retain actor-specific sampling and objectives; full run settings are in the repository.}
\label{tab:implementation}
\begin{tabular}{lcl}
\toprule
Method & Actor steps & Policy objective or density\\
\midrule
SAC & --- & Gaussian log-density\\
SAC-Flow & 4 & Joint-path entropy\\
DIME & 16 & Diffusion entropy bound\\
QSM & 5 & Q-score matching (no log-density)\\
DEFlow & 4 & Discrete change-of-variables density\\
\midrule
\multicolumn{3}{l}{Shared: twin $4\times512$ critics; batch 512; replay $10^6$; warm-up 50k; UTD 1.}\\
\multicolumn{3}{l}{$\gamma=0.99$; $\tau=0.005$; actor/critic learning rates $3\times10^{-4}/10^{-3}$.}\\
\bottomrule
\end{tabular}
\end{table}

\subsection{Benchmark Protocol}
\label{app:experimental-setup}

The benchmarks comprise MultiGoal-4 \citep{haarnoja2017sql}, five MuJoCo locomotion tasks (Hopper-v4, Walker2d-v4, HalfCheetah-v4, Ant-v4, Humanoid-v4) \citep{todorov2012mujoco}, two ManiSkill tasks (PickCube-v1 and PushCube-v1) \citep{tao2025maniskill3}, and three MetaWorld tasks (ButtonPressWall, PegInsertSide, PushWall) \citep{yu2020metaworld}. All methods use the same task wrapper and checkpoint schedule within each comparison. MuJoCo and manipulation runs use budgets of 1M and 500k recorded steps, respectively. Manipulation wrappers set task-specific rewards, resets, horizons, and termination rules; their exact settings are provided in the repository's run configurations.

MuJoCo returns use the recorded evaluation at 1M steps, curves use evaluations every 25k steps. Manipulation returns are evaluated from 50k to 500k steps and averaged over the recorded checkpoints for the time-averaged-return table. We report means and sample standard deviations across three training seeds per task and method. For deterministic DEFlow evaluation, the base sample is $z=0$, QSM retains stochastic denoising even in its evaluation channel. Final manipulation ever-success is the fraction of sampled episodes in which the task succeeds at least once, measured over 100 episodes per trained policy. These evaluation episodes are not additional training seeds. The exact selected runs and evaluation scripts are linked from the repository's result manifests.

\subsection{Fixed-Flow Robot Illustration}
\label{app:fixed-flow-robot}

Figure~\ref{fig:n2-robot} isolates discretization in a fixed scalar contraction,
$v(x,t)=-8(x-0.8)$, on $t\in[0,1]$, initialized with $z\sim\mathcal N(0,1)$.
We compare explicit Euler with a jointly solved implicit-Euler trajectory at
$T\in\{2,3,4,5\}$ and apply $u=\tanh(x_T)$ to the terminal sample.
The implicit solver uses damped Newton with zero initialization, an
80-iteration cap, and normalized local and cumulative fixed-point residual
tolerances of $10^{-11}$ and $10^{-9}$, respectively, in double precision.
This diagnostic has no learned velocity field, critic, or policy training;
the learned DEFlow policies in Sec.~\ref{sec:method} use Anderson acceleration.

The simulator is a custom one-decision Panda waypoint-reaching task in
ManiSkill~3. The scalar command specifies a displacement of $0.12u$ metres
along the task axis from the calibrated end-effector anchor. The target uses
$u_\star=\tanh(0.8(1-e^{-8}))$, the continuous-flow image of $z=0$ after
squashing. Each waypoint is held for 40 Cartesian position-control steps at
20~Hz, with the gripper fixed and no early success termination. Success is
the final three-dimensional end-effector distance to the target being less
than $8$~mm; it is not an ever-success metric. The plot shows axial and
lateral projections, while its success labels use all three coordinates.

Each method and step count uses 640 trials: 128 trials for each of seeds
$0,1,2,3,4$. Simulator states and Gaussian base samples are shared across
methods and step counts. The figure reuses recorded endpoints from the
descriptive step-count sweep and is not an independent replication.
The comparison holds the number of integration steps fixed, without matching
field evaluations or wall-clock time. It supports a discretization comparison
for this contraction task, without establishing computational efficiency,
multimodal sampling accuracy, or performance of a trained policy.

\section{Additional Results}
\label{app:results}

\subsection{Numerical Verification}
\label{sec:verify}
\label{app:numerical-implementation}

DEFlow solves the joint trajectory with Anderson acceleration \citep{walker2011anderson}. The solver stops when the minibatch mean relative residual meets its configured tolerance or the iteration limit is reached, returning the last iterate in either case. Actor updates re-evaluate the field at that iterate and use a dense adjoint solve rather than differentiating through solver iterations. The adjoint and log-determinant matrices receive a $10^{-6}I$ shift, and log-determinants are clipped to $[-50,50]$. Some training recipes also clip actor log-densities to $[-15,15]$. The reference MuJoCo configuration uses a $3\times10^{-4}$ forward tolerance.

We test how the forward solver affects trained DEFlow policies by comparing each final manipulation policy under two settings: the original Anderson solver with a $25$-iteration limit and $10^{-5}$ tolerance, and a tighter solver with a $100$-iteration limit and $10^{-6}$ tolerance. Table~\ref{tab:verify} compares actions, safeguarded log-density readouts, and actor gradients with an independently solved implicit reference; the median and P95 pool matched inputs within each task. Three PushWall inputs without a validated reference are excluded from these discrepancies. Table~\ref{tab:verify-success} reports stochastic ever-success on paired environment episodes as the mean $\pm$ sample standard deviation over three training seeds. Both settings use the same $100$ evaluation episodes per checkpoint, and the change is Tighter minus Original in percentage points.

\begin{center}
\begin{minipage}{\textwidth}
\centering\small\setlength{\tabcolsep}{3.5pt}
\captionof{table}{Numerical discrepancies on trained policies.}
\label{tab:verify}
\begin{tabular}{llrrrr}
\toprule
& & \multicolumn{2}{c}{Original} & \multicolumn{2}{c}{Tighter}\\
\cmidrule(lr){3-4}\cmidrule(lr){5-6}
Task & Discrepancy & Median & P95 & Median & P95\\
\midrule
PickCube & Action & $1.49{\times}10^{-5}$ & $1.62{\times}10^{-4}$ & $7.04{\times}10^{-7}$ & $2.39{\times}10^{-6}$\\
& Log-density (nat) & $3.73{\times}10^{-5}$ & $5.18{\times}10^{-4}$ & $2.05{\times}10^{-6}$ & $9.92{\times}10^{-6}$\\
& Gradient (\%) & $0.00330$ & $0.771$ & $0.000143$ & $0.000980$\\
\midrule
PushCube & Action & $1.02{\times}10^{-5}$ & $3.20{\times}10^{-5}$ & $9.73{\times}10^{-7}$ & $2.38{\times}10^{-6}$\\
& Log-density (nat) & $2.03{\times}10^{-5}$ & $8.15{\times}10^{-5}$ & $2.47{\times}10^{-6}$ & $8.91{\times}10^{-6}$\\
& Gradient (\%) & $0.000930$ & $0.00465$ & $0.000229$ & $0.000642$\\
\midrule
ButtonPressWall & Action & $1.63{\times}10^{-5}$ & $9.92{\times}10^{-4}$ & $9.95{\times}10^{-7}$ & $3.10{\times}10^{-6}$\\
& Log-density (nat) & $4.49{\times}10^{-5}$ & $1.91{\times}10^{-3}$ & $2.55{\times}10^{-6}$ & $1.05{\times}10^{-5}$\\
& Gradient (\%) & $0.00630$ & $0.836$ & $0.000420$ & $0.00178$\\
\midrule
PegInsertSide & Action & $8.54{\times}10^{-6}$ & $2.75{\times}10^{-5}$ & $9.69{\times}10^{-7}$ & $2.57{\times}10^{-6}$\\
& Log-density (nat) & $2.22{\times}10^{-5}$ & $1.31{\times}10^{-4}$ & $2.09{\times}10^{-6}$ & $9.18{\times}10^{-6}$\\
& Gradient (\%) & $0.000523$ & $0.00435$ & $0.000153$ & $0.000390$\\
\midrule
PushWall & Action & $0.00299$ & $0.0375$ & $0.000479$ & $0.0211$\\
& Log-density (nat) & $0.0108$ & $0.299$ & $0.00122$ & $0.197$\\
& Gradient (\%) & $13.7$ & $96.1$ & $2.84$ & $50.3$\\
\bottomrule
\end{tabular}
\end{minipage}
\end{center}

\begin{center}
\begin{minipage}{\textwidth}
\centering\small\setlength{\tabcolsep}{5pt}
\captionof{table}{Final sampled success under each solver.}
\label{tab:verify-success}
\begin{tabular}{lrrr}
\toprule
Task & Original & Tighter & Change (pp)\\
\midrule
PickCube & $87.0\pm8.9$ & $86.7\pm8.7$ & $-0.33$\\
PushCube & $98.7\pm1.2$ & $99.0\pm1.0$ & $+0.33$\\
ButtonPressWall & $8.7\pm14.2$ & $8.3\pm13.6$ & $-0.33$\\
PegInsertSide & $38.0\pm23.8$ & $38.0\pm23.8$ & $0.00$\\
PushWall & $0.3\pm0.6$ & $0.3\pm0.6$ & $0.00$\\
\bottomrule
\end{tabular}
\end{minipage}
\end{center}

The tighter settings reduce the median and P95 discrepancies for every task and quantity in Table~\ref{tab:verify}. Final sampled success changes little on the paired episodes in Table~\ref{tab:verify-success}. This means that a more accurate solver at evaluation does not improve task success overall. For the iteration limit, the fraction of evaluation calls that reach it averages $47.2\%$ with Original and $13.1\%$ with Tighter across checkpoints. These results measure the sensitivity of final policies to the tested solver change, while errors during learning and the likelihood of the finite-iteration sampler remain open. Original is reevaluated in the same runtime as Tighter and can differ from the earlier benchmark evaluation.

\subsection{Extended Benchmark Results}
\label{app:tables}

\begin{center}
\begin{minipage}{\textwidth}
\centering
\includegraphics[width=\textwidth]{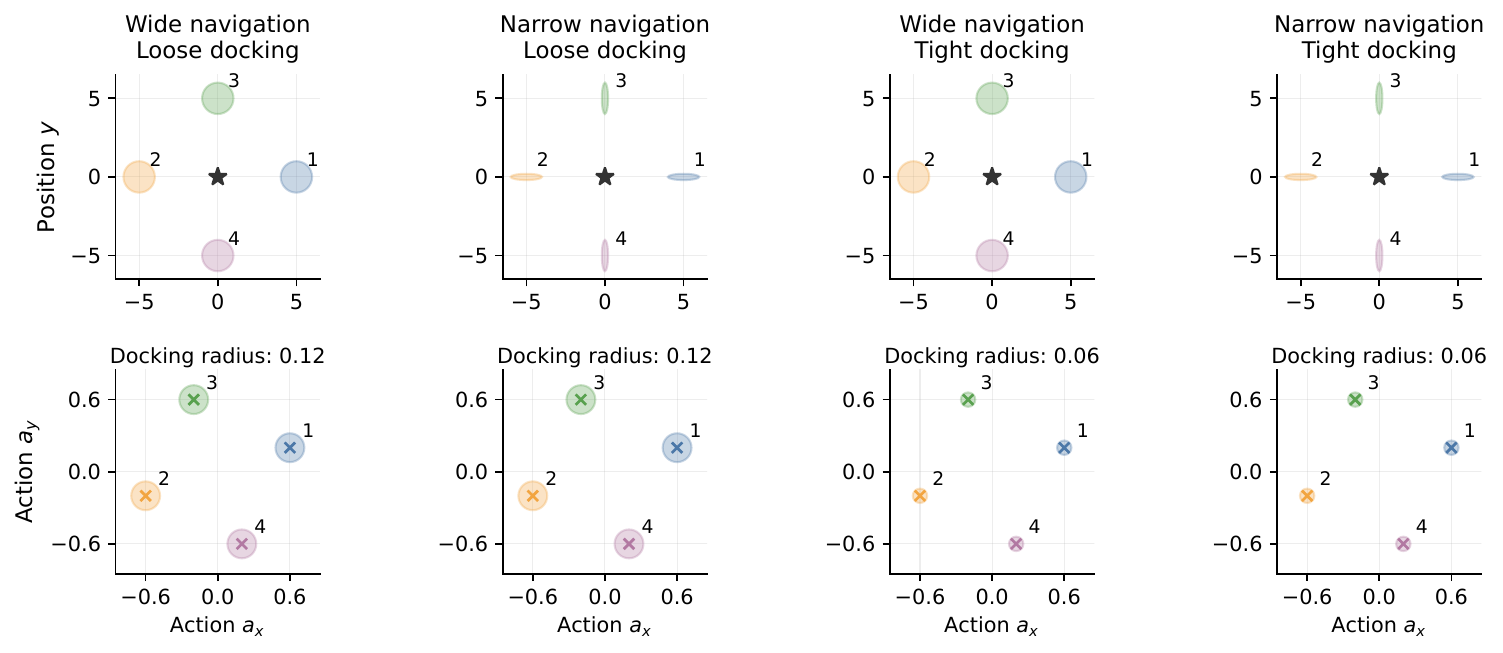}
\captionof{figure}{Four MultiGoal docking variants. Top: navigation goal regions. Bottom: corresponding docking targets and action tolerances; colors and numbers identify goals.}
\label{fig:multigoal}
\end{minipage}
\end{center}

\begin{center}
\begin{minipage}{\textwidth}
\centering
\includegraphics[width=\textwidth]{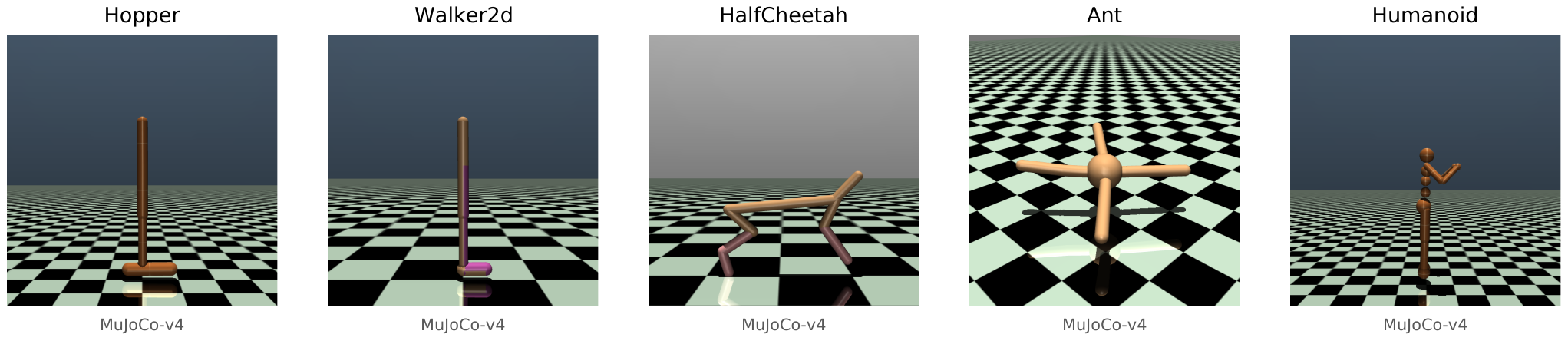}
\captionof{figure}{MuJoCo continuous-control environments used in our experiments.}
\label{fig:mujoco-environments}
\end{minipage}
\end{center}

\begin{center}
\begin{minipage}{\textwidth}
\centering
\includegraphics[width=\textwidth]{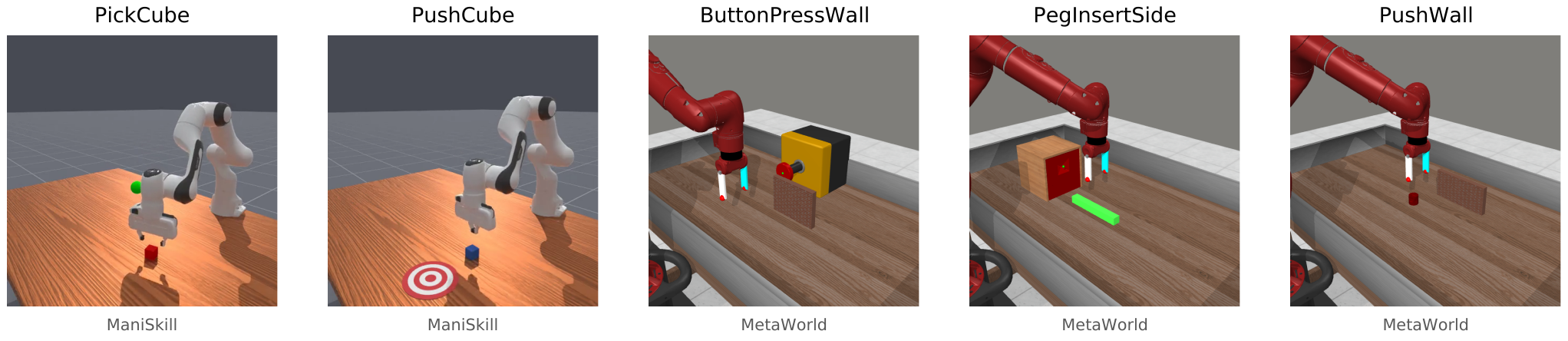}
\captionof{figure}{Manipulation tasks from ManiSkill (first two) and MetaWorld (last three). ManiSkill images are from the official task demonstrations.}
\label{fig:manipulation-environments}
\end{minipage}
\end{center}

\begin{center}
\begin{minipage}{\textwidth}
\centering\small\setlength{\tabcolsep}{4pt}
\captionof{table}{Final sampled ever-success (\%) at 500k steps, mean $\pm$ SD across three training seeds. Bold and underlined values mark the top two means.}
\label{tab:manip}
\begin{tabular*}{\textwidth}{@{\extracolsep{\fill}}lrrrr@{}}
\toprule
Task & SAC & SAC-Flow & DIME & DEFlow \\
\midrule
PickCube & $64.7 \pm 56.0$ & $5.0 \pm 8.7$ & $\mathbf{93.0 \pm 9.5}$ & $\underline{87.3 \pm 8.3}$ \\
PushCube & $97.0 \pm 3.6$ & $97.3 \pm 3.1$ & $\mathbf{99.0 \pm 1.7}$ & $\underline{98.7 \pm 1.2}$ \\
ButtonPressWall & $0.0 \pm 0.0$ & $0.0 \pm 0.0$ & $\underline{3.7 \pm 2.1}$ & $\mathbf{8.7 \pm 14.2}$ \\
PegInsertSide & $\mathbf{72.3 \pm 9.8}$ & $0.0 \pm 0.0$ & $23.7 \pm 41.0$ & $\underline{39.3 \pm 21.5}$ \\
PushWall & $0.0 \pm 0.0$ & $\mathbf{10.3 \pm 17.9}$ & $\underline{1.0 \pm 1.7}$ & $0.3 \pm 0.6$ \\
\midrule
Sum & $\underline{234.0}$ & $112.7$ & $220.3$ & $\mathbf{234.3}$ \\
\bottomrule
\end{tabular*}

\end{minipage}
\end{center}

\begin{table}[htbp]
\centering\small\setlength{\tabcolsep}{4pt}
\caption{MuJoCo return at 1M steps, mean $\pm$ SD across three runs. Bold and underlined values mark the highest and second-highest means.}
\label{tab:mean-return}
\begin{tabular}{lrrrrr}
\toprule
Environment & SAC & SAC-Flow & DIME & QSM & DEFlow \\
\midrule
Hopper & $2510 \pm 901$ & $\underline{2955 \pm 1055}$ & $2485 \pm 326$ & $2474 \pm 1284$ & $\mathbf{3318 \pm 560}$ \\
Walker2d & $4517 \pm 644$ & $4513 \pm 436$ & $\underline{4866 \pm 348}$ & $1615 \pm 992$ & $\mathbf{4973 \pm 564}$ \\
HalfCheetah & $\mathbf{14190 \pm 1152}$ & $11476 \pm 1647$ & $\underline{13476 \pm 527}$ & $5981 \pm 791$ & $12534 \pm 340$ \\
Ant & $\underline{5995 \pm 188}$ & $\mathbf{6215 \pm 458}$ & $5975 \pm 568$ & $-22 \pm 2$ & $5673 \pm 1340$ \\
Humanoid & $\underline{5042 \pm 334}$ & $\mathbf{5089 \pm 123}$ & $4981 \pm 185$ & $4672 \pm 191$ & $4211 \pm 858$ \\
\midrule
Sum & $\mathbf{32253}$ & $30247$ & $\underline{31783}$ & $14721$ & $30709$ \\
\bottomrule
\end{tabular}

\end{table}

\begin{table}[htbp]
\centering\small\setlength{\tabcolsep}{4pt}
\caption{Time-averaged return over 50k--500k steps, mean $\pm$ SD across three seeds. Bold and underlined values mark the top two means.}
\label{tab:manip-return}
\begin{tabular}{lrrrr}
\toprule
Environment & SAC & SAC-Flow & DIME & DEFlow \\
\midrule
MS PickCube & $25.1 \pm 5.9$ & $19.0 \pm 1.5$ & $\underline{29.1 \pm 3.0}$ & $\mathbf{33.5 \pm 0.5}$ \\
MS PushCube & $\underline{38.4 \pm 0.7}$ & $37.3 \pm 1.2$ & $37.3 \pm 1.5$ & $\mathbf{40.8 \pm 0.2}$ \\
MW ButtonPressWall & $641.6 \pm 91.1$ & $47.2 \pm 55.6$ & $\underline{696.7 \pm 30.1}$ & $\mathbf{755.7 \pm 24.9}$ \\
MW PegInsertSide & $\mathbf{699.1 \pm 209.1}$ & $2.5 \pm 0.0$ & $298.2 \pm 237.4$ & $\underline{564.3 \pm 158.7}$ \\
MW PushWall & $\underline{61.2 \pm 39.3}$ & $28.1 \pm 38.5$ & $23.4 \pm 19.4$ & $\mathbf{217.8 \pm 334.9}$ \\
\midrule
Sum & $\underline{1465.4}$ & $134.0$ & $1084.9$ & $\mathbf{1612.2}$ \\
\bottomrule
\end{tabular}

\end{table}

Table~\ref{tab:mean-return} reports MuJoCo return at the recorded 1M evaluation.
Table~\ref{tab:manip-return} reports time-averaged manipulation return over 50k--500k steps.
Table~\ref{tab:return-sensitivity} reports the last-100k mean, return AUC, and return at 950k for the MuJoCo benchmark.
Table~\ref{tab:manip-runs} reports return at 500k, return AUC over 50k--500k steps, and sampled ever-success for the manipulation benchmark.

\begin{center}
\begin{minipage}{\textwidth}
\centering\scriptsize\setlength{\tabcolsep}{6pt}
\captionof{table}{Sensitivity analysis: mean $\pm$ 95\% confidence-interval half-width across three training seeds per cell. Return AUC is the time-averaged return over 25k--1M steps.}
\label{tab:return-sensitivity}
\begin{tabular*}{\textwidth}{@{\extracolsep{\fill}}lrrrrr@{}}
\toprule
Environment & SAC & SAC-Flow & DIME & QSM & DEFlow \\
\midrule
\multicolumn{6}{@{}l}{\textit{Last 100k mean}} \\
Hopper & $2775 \pm 644$ & $3328 \pm 591$ & $2416 \pm 627$ & $2491 \pm 1277$ & $3165 \pm 1126$ \\
Walker2d & $4487 \pm 1604$ & $4443 \pm 844$ & $4849 \pm 964$ & $1461 \pm 2336$ & $5169 \pm 522$ \\
HalfCheetah & $14098 \pm 2417$ & $11533 \pm 3924$ & $13501 \pm 1278$ & $5982 \pm 999$ & $12723 \pm 449$ \\
Ant & $5737 \pm 859$ & $6286 \pm 530$ & $6047 \pm 1171$ & $-25 \pm 6$ & $5824 \pm 2296$ \\
Humanoid & $4345 \pm 1267$ & $5023 \pm 107$ & $4875 \pm 512$ & $4539 \pm 1046$ & $4357 \pm 1914$ \\
\midrule
\multicolumn{6}{@{}l}{\textit{Return AUC}} \\
Hopper & $2157 \pm 1123$ & $2644 \pm 583$ & $2500 \pm 164$ & $1956 \pm 762$ & $2858 \pm 567$ \\
Walker2d & $2873 \pm 2416$ & $3112 \pm 176$ & $3127 \pm 1230$ & $879 \pm 1219$ & $3583 \pm 628$ \\
HalfCheetah & $11220 \pm 999$ & $9806 \pm 2582$ & $11564 \pm 461$ & $5237 \pm 610$ & $10465 \pm 1491$ \\
Ant & $3569 \pm 715$ & $4917 \pm 846$ & $4470 \pm 1235$ & $-27 \pm 4$ & $3846 \pm 651$ \\
Humanoid & $2902 \pm 462$ & $3400 \pm 282$ & $3150 \pm 816$ & $2722 \pm 559$ & $2706 \pm 496$ \\
\midrule
\multicolumn{6}{@{}l}{\textit{Return at 950k}} \\
Hopper & $3543 \pm 828$ & $3559 \pm 45$ & $2251 \pm 2148$ & $2996 \pm 1308$ & $2725 \pm 1920$ \\
Walker2d & $4423 \pm 1504$ & $3748 \pm 2373$ & $4757 \pm 1307$ & $1035 \pm 1551$ & $5131 \pm 512$ \\
HalfCheetah & $13709 \pm 2787$ & $11624 \pm 3846$ & $13071 \pm 3973$ & $6196 \pm 917$ & $12677 \pm 560$ \\
Ant & $5682 \pm 973$ & $6145 \pm 1255$ & $6046 \pm 1664$ & $-25 \pm 2$ & $5583 \pm 2774$ \\
Humanoid & $4441 \pm 1126$ & $4848 \pm 743$ & $5220 \pm 459$ & $4407 \pm 2007$ & $4250 \pm 3408$ \\
\bottomrule
\end{tabular*}

\end{minipage}
\end{center}

\begin{center}
\begin{minipage}{\textwidth}
\centering\scriptsize\setlength{\tabcolsep}{6pt}
\captionof{table}{Manipulation benchmark results, mean $\pm$ 95\% confidence-interval half-width across three training seeds per cell. Return AUC is the time-averaged return over 50k--500k steps.}
\label{tab:manip-runs}
\begin{tabular*}{\textwidth}{@{\extracolsep{\fill}}lrrrr@{}}
\toprule
Environment & SAC & SAC-Flow & DIME & DEFlow \\
\midrule
\multicolumn{5}{@{}l}{\textit{Return at 500k}} \\
PickCube & $30.2 \pm 28.3$ & $23.2 \pm 6.6$ & $38.7 \pm 7.1$ & $35.5 \pm 3.7$ \\
PushCube & $40.3 \pm 11.2$ & $39.8 \pm 10.5$ & $42.7 \pm 1.0$ & $42.8 \pm 1.2$ \\
ButtonPressWall & $845.6 \pm 108.9$ & $35.9 \pm 89.1$ & $772.8 \pm 155.2$ & $747.6 \pm 200.3$ \\
PegInsertSide & $1274.5 \pm 308.7$ & $2.0 \pm 0.5$ & $699.7 \pm 1607.9$ & $1091.7 \pm 693.2$ \\
PushWall & $78.7 \pm 118.1$ & $7.7 \pm 19.0$ & $15.8 \pm 12.5$ & $457.8 \pm 1887.2$ \\
\midrule
\multicolumn{5}{@{}l}{\textit{Return AUC}} \\
PickCube & $25.1 \pm 14.6$ & $19.0 \pm 3.7$ & $29.1 \pm 7.5$ & $33.5 \pm 1.2$ \\
PushCube & $38.4 \pm 1.8$ & $37.3 \pm 3.1$ & $37.3 \pm 3.7$ & $40.8 \pm 0.6$ \\
ButtonPressWall & $641.6 \pm 226.3$ & $47.2 \pm 138.2$ & $696.7 \pm 74.8$ & $755.7 \pm 61.8$ \\
PegInsertSide & $699.1 \pm 519.5$ & $2.5 \pm 0.1$ & $298.2 \pm 589.7$ & $564.3 \pm 394.2$ \\
PushWall & $61.2 \pm 97.6$ & $28.1 \pm 95.6$ & $23.4 \pm 48.3$ & $217.8 \pm 831.9$ \\
\midrule
\multicolumn{5}{@{}l}{\textit{Sampled ever-success (\%)}} \\
PickCube & $64.7 \pm 139.1$ & $5.0 \pm 21.5$ & $93.0 \pm 23.7$ & $87.3 \pm 20.7$ \\
PushCube & $97.0 \pm 9.0$ & $97.3 \pm 7.6$ & $99.0 \pm 4.3$ & $98.7 \pm 2.9$ \\
ButtonPressWall & $0.0 \pm 0.0$ & $0.0 \pm 0.0$ & $3.7 \pm 5.2$ & $8.7 \pm 35.2$ \\
PegInsertSide & $72.3 \pm 24.4$ & $0.0 \pm 0.0$ & $23.7 \pm 101.8$ & $39.3 \pm 53.3$ \\
PushWall & $0.0 \pm 0.0$ & $10.3 \pm 44.5$ & $1.0 \pm 4.3$ & $0.3 \pm 1.4$ \\
\bottomrule
\end{tabular*}
\end{minipage}
\end{center}

\subsection{Ablation Studies}
\label{app:kabl-curves}

\paragraph{Integrator step-count ablation.}
Both figures use the wide-goal, tight-docking MultiGoal variant and two frozen policies, each keeping its training integrator (explicit Euler, implicit DEFlow), both trained at $T=4$ (seed 3, 200005 transitions). We vary only the docking step count $T\in\{2,4,8,16\}$, keeping navigation and all inputs fixed; this is the only seed with both final checkpoints, since explicit seeds 1 and 2 failed numerically. Figure~\ref{fig:kabl-shared} evaluates both policies on the same 104 reached states with 16 Gaussian inputs each (1664 actions), using balanced Hit@0.06, the per-goal hit rates averaged with equal weights. Figure~\ref{fig:kabl-own} follows 1000 new stochastic $T=4$ episodes per policy and recomputes only the docking action from each recorded state, Gaussian base sample, and full batch; replaying at $T=4$ reproduces the recorded actions bit-for-bit and matches every recorded episode success. Its Hit is the fraction of the 1000 docking actions within 0.06 of the reached goal's target, so frequent routes dominate.

On the shared states, DEFlow stays high for $T\ge4$ (94.1\%, 96.5\%, 95.1\%; 54.9\% at $T=2$), while the explicit policy reaches 50.1\% at its training depth, accurate on only two goals and near zero on the other two, and degrades at smaller and larger step counts (1.4\%, 35.1\%, 15.1\% at $T=2,8,16$). On its own trajectories, explicit succeeds on 99.8\% of episodes at $T=4$ because they visit almost only those two goals, then drops to 0.7\% at $T=2$ and 59.1\%/28.2\% at $T=8$/$16$; DEFlow holds 96.5\%, 99.1\%, 98.0\%, and 53.7\% at $T=2$. In this pair, DEFlow's docking is nearly insensitive to the tested step counts, whereas the explicit policy's accuracy is route-selective and strongly step-dependent (paired illustration, one seed, each policy with its own integrator).

\begin{center}
\begin{minipage}{\textwidth}
\centering
\includegraphics[width=\textwidth]{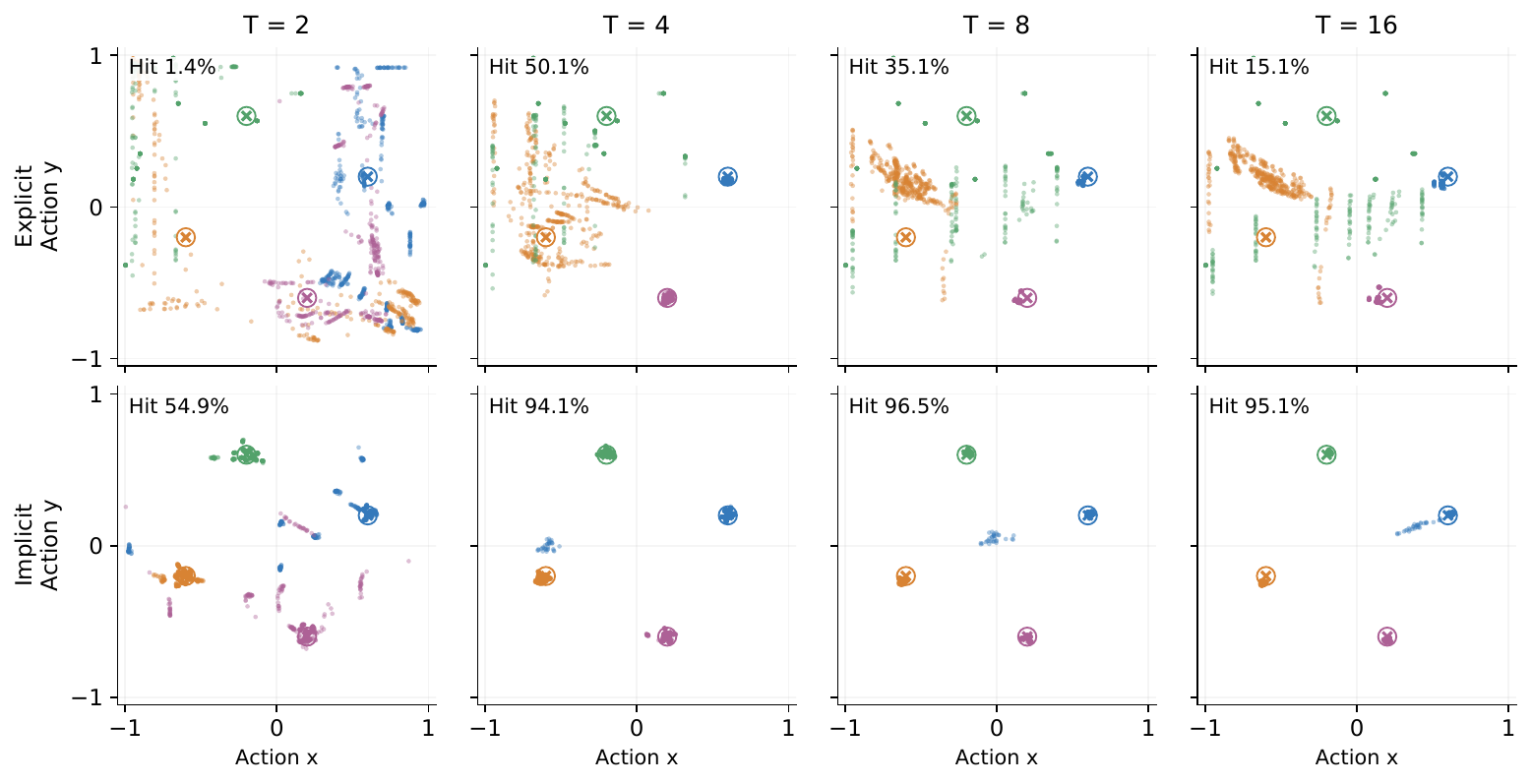}
\captionof{figure}{Docking actions on shared states as $T$ varies. Top: explicit; bottom: DEFlow. Hit averages equally over goals.}
\label{fig:kabl-shared}
\end{minipage}
\end{center}

\begin{center}
\begin{minipage}{\textwidth}
\centering
\includegraphics[width=\textwidth]{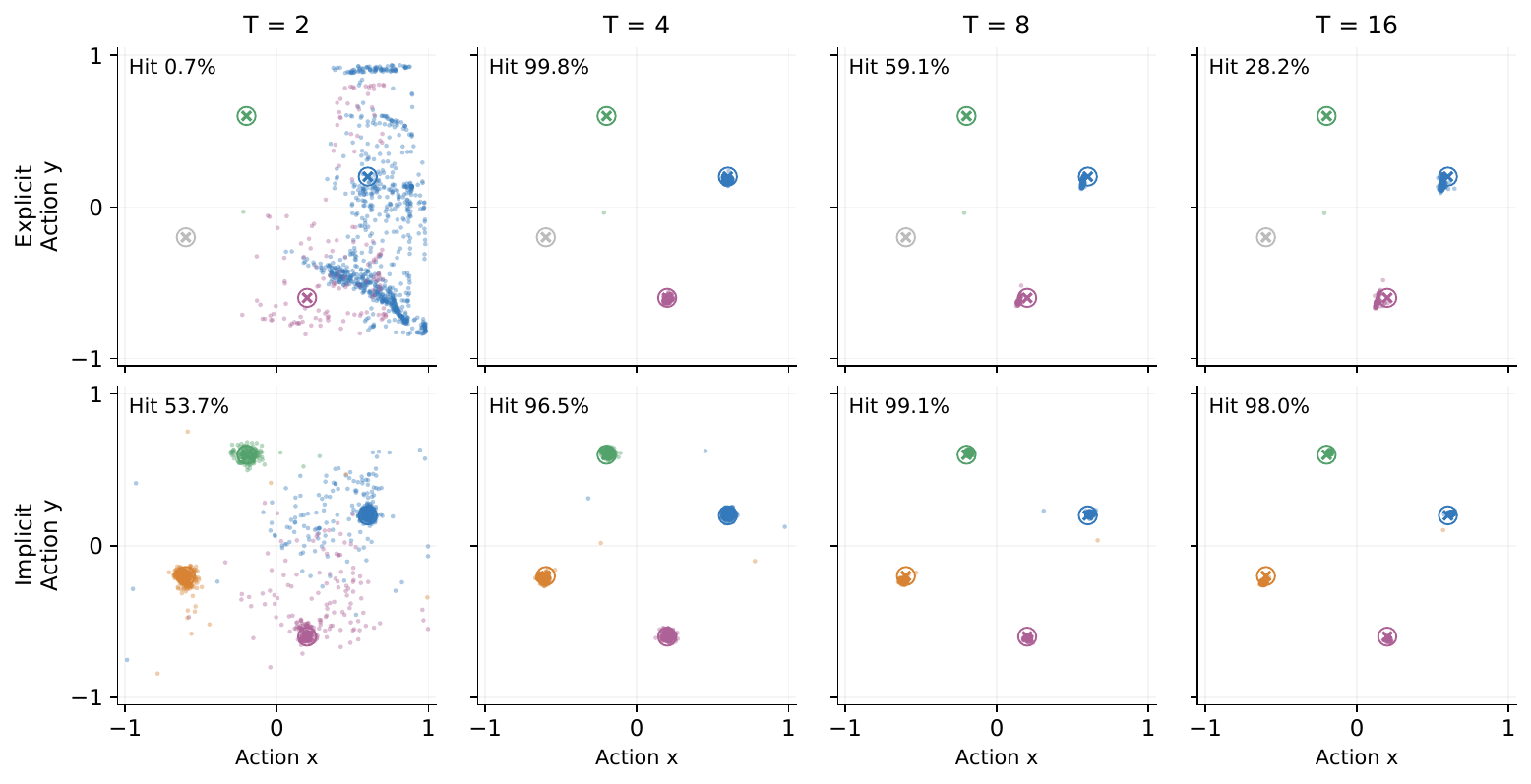}
\captionof{figure}{Docking actions on each policy's own trajectories. Hit uses observed goal frequencies; gray targets were not visited.}
\label{fig:kabl-own}
\end{minipage}
\end{center}

\clearpage
\subsection{Docking Failure Diagnostics}
\label{app:docking-failures}

\paragraph{Explicit flow.}
The docking audit illustrates the step-map conditioning effect identified in
Sec.~\ref{sec:problem}. In the notation of
Eq.~\eqref{eq:problem-step-jacobians}, the explicit-step Jacobian is
$B_k=D_aF_k^{\mathrm{exp}}=I+\Delta t J_k$. Three original explicit-flow
runs terminated after detecting non-finite TD targets. At a frozen checkpoint
from one of these runs, $\det(B_0+\varepsilon I)$ changes sign between two
sampled latent inputs. A float64 probe locates a singularity of this
diagonally regularized matrix between them. At the same local $J_0$, the
algebraic implicit counterpart $I-\Delta t J_0$ has minimum singular value
$1.226$. The checkpoint thus exhibits the singular explicit-step regime
illustrated by the linear contraction in
Appendix~\ref{app:stabilization-counterexamples}.

\paragraph{QSM.}
QSM's zero docking success is not a failure to reach the goal: 1{,}374 of 3{,}000 episodes across three seeds reached the docking state, yet none docked. The failure lies in action generation. At a frozen checkpoint, ideal goal-center observations produced action-box corners in 2{,}045 of 2{,}048 samples. An exact replay of the five denoising steps shows that the final learned correction has median magnitude 16.0--21.6 in a box of $[-1,1]^2$, pushing both coordinates outside the box in 99.4--100\% of samples, so clipping maps the action to a corner. All values are finite and the final step adds no fresh noise. Why the correction is learned so large remains unresolved.

\section{AI Assistance}

We used AI tools to assist with writing and polishing the manuscript, to support literature retrieval and discovery of related work, and to draft sections of the paper. The authors reviewed and revised the resulting text and take full responsibility for the final manuscript.

\end{document}